\documentclass{article}

    \usepackage[final, nonatbib]{neurips_2023}

\usepackage[utf8]{inputenc} 
\usepackage[T1]{fontenc}    
\usepackage{hyperref}       
\usepackage{url}            
\usepackage{booktabs}       
\usepackage{amsfonts}       
\usepackage{nicefrac}       
\usepackage{microtype}      
\usepackage{xcolor}         

\usepackage{mathtools}
\usepackage{algorithm}
\usepackage{algpseudocode}
\usepackage{amsmath,amssymb,amsthm,amsfonts}
\usepackage{subfigure}
\usepackage{url}
\usepackage{makecell}
\usepackage{multirow}
\usepackage[symbol]{footmisc}
\usepackage{tikz}
\usepackage{caption}
\usepackage{hyperref}

\usetikzlibrary{shapes,decorations,arrows,calc,arrows.meta,fit,positioning}
\tikzset{
    -Latex,auto,node distance =1 cm and 1 cm,semithick,
    state/.style ={ellipse, draw, minimum width = 0.7 cm},
    point/.style = {circle, draw, inner sep=0.04cm,fill,node contents={}},
    bidirected/.style={Latex-Latex,dashed},
    el/.style = {inner sep=2pt, align=left, sloped}
}

\newcommand{\bz}{{\bf z}}

\newcommand{\by}{{\bf y}}

\newcommand{\bx}{{\bf x}}

\newcommand{\mbR}{\mathbb{R}}
\newcommand{\mbE}{\mathbb{E}}

\newcommand{\bc}{\begin{center}}
\newcommand{\ec}{\end{center}}
\newcommand{\be}{\begin{equation}}
\newcommand{\ee}{\end{equation}}
\newcommand{\ba}{\begin{array}}
\newcommand{\ea}{\end{array}}
\newcommand{\bean}{\begin{eqnarray*}}
\newcommand{\eean}{\end{eqnarray*}}
\newcommand{\bea}{\begin{eqnarray}}
\newcommand{\eea}{\end{eqnarray}}
\newcommand{\ben}{\begin{enumerate}}
\newcommand{\een}{\end{enumerate}}
\newcommand{\bed}{\begin{itemize}}
\newcommand{\eed}{\end{itemize}}
\newcommand{\bs}{\begin{slide}}
\newcommand{\es}{\end{slide}}

\newtheorem{theorem}{Theorem}
\newtheorem*{theorem*}{Theorem}

\newtheorem*{prop*}{Proposition}

\newtheorem{assumption}{Assumption}

\title{Distributional Learning of Variational AutoEncoder: Application to Synthetic Data Generation}

\author{
  Seunghwan An \textnormal{and} Jong-June Jeon\thanks{Corresponding author.} \\
  Department of Statistical Data Science, University of Seoul, S. Korea \\
  \texttt{\{dkstmdghks79, jj.jeon\}@uos.ac.kr}
}

\begin{document}

\maketitle

\begin{abstract}
The Gaussianity assumption has been consistently criticized as a main limitation of the Variational Autoencoder (VAE) despite its efficiency in computational modeling. In this paper, we propose a new approach that expands the model capacity (i.e., expressive power of distributional family) without sacrificing the computational advantages of the VAE framework. Our VAE model's decoder is composed of an infinite mixture of asymmetric Laplace distribution, which possesses general distribution fitting capabilities for continuous variables. Our model is represented by a special form of a nonparametric M-estimator for estimating general quantile functions, and we theoretically establish the relevance between the proposed model and quantile estimation. We apply the proposed model to synthetic data generation, and particularly, our model demonstrates superiority in easily adjusting the level of data privacy.
\end{abstract}

\section{Introduction}
\label{sec:1}

Variational Autoencoder (VAE) \cite{Kingma2014, JimenezRezende2014StochasticBA} and Generative Adversarial Networks (GAN) \cite{Goodfellow2014GenerativeAN} are generative models that are used to estimate the underlying distribution of a given dataset. To avoid the curse of dimensionality, VAE and GAN commonly introduce a low-dimensional latent space on which a conditional generative model is defined. By minimizing an information divergence between the original data and its generated data, the generative models are learned to produce synthetic data similar to the original one. Accordingly, VAE and GAN have been applied in various applications, such as generating realistic images, texts, and synthetic tabular data for privacy preservation purposes \cite{Karras2018ASG, wang-etal-2019-topic, NEURIPS2019_254ed7d2, pmlr-v157-zhao21a, kiran2023comparative}.

However, the difference in the strength of the assumption about the generative distribution brings significant contrasts in the VAE and GAN generation performances \cite{Karras2018ASG, brock2018large, DiazPinto2018RetinalIS}. In the GAN framework, the adversarial loss enables direct minimization of the Jensen-Shannon divergence between the ground-truth density function and the generative distribution under no distributional assumption \cite{creswell2018generative, wang2017generative}. Roughly speaking, the GAN employs a nonparametric model as its conditional generative model defined on the latent space.

On the contrary, in the VAE framework, the Gaussianity assumption has been favored \cite{Kingma2014, kingma2014semi, dai2018diagnosing, Castrejn2019ImprovedCV, NEURIPS2020_08058bf5}. It is because Gaussianity gives us three advantages: 1) the reconstruction loss can be interpreted as the mean squared error that is one of the most popular losses in optimization theory, 2) generating a new sample is computationally straightforward, and 3) KL-divergence is computed in a simple closed form. However, these benefits have led us to pay the price for the distributional capacity of the generative model, in that the generative model of the VAE is constrained in the form of marginalization of the product of the two Gaussian distributions. Here, the distributional capacity means the expressive power of the distributional family. This restricted distributional capacity has been the critical limitation \cite{Burda2015ImportanceWA, NIPS2016_ddeebdee} and leads to a heavy parameterization of the decoder mean vector to approximate complex underlying distributions. 

To increase the distributional capacity in synthetic data generation, \cite{NEURIPS2019_254ed7d2, pmlr-v157-zhao21a} introduce the multi-modality in the distributional assumption of the decoder, which is known as the \textit{mode-specific normalization technique}. Although the mixture Gaussian decoder modeling of \cite{NEURIPS2019_254ed7d2, pmlr-v157-zhao21a} allows handling more complex distributions of the observed dataset while preserving all of the advantages of Gaussianity, we numerically find that the mixture Gaussian is not enough to capture the complex underlying distribution.

Our main contribution is that, beyond Gaussianity, we propose a novel VAE learning method that directly estimates the conditional cumulative distribution function (CDF) while maintaining the objective of maximizing the Evidence Lower Bound (ELBO) of the observed dataset. It implies that we have a nonparametric distribution assumption on the generative model. We call this approach \textit{distributional learning of the VAE}, which is enabled by estimating an infinite number of conditional quantiles \cite{brando2019modelling, gasthaus2019probabilistic}. By adopting the \textit{continuous ranked probability score} (CRPS) loss, the objective function of our proposed distribution learning method is computationally tractable \cite{Gneiting2007StrictlyPS, Matheson1976ScoringRF, gasthaus2019probabilistic}. 

In our proposed distributional learning framework, 1) the reconstruction loss is equivalent to the CRPS loss, which is a \textit{proper scoring rule} \cite{Gneiting2007StrictlyPS, Matheson1976ScoringRF}, 2) generating a new sample is still computationally straightforward due to the inverse transform sampling, and 3) KL-divergence is still computed in a simple closed form. To show the effectiveness of our proposed model in capturing the underlying distribution of the dataset, we evaluate our model for synthetic data generation with real tabular datasets. 

\section{Related Work}
\label{sec:2}

\textbf{Modeling of the decoder and reconstruction loss.} 
To increase the distributional capacity, many papers have focused on decoder modeling while not losing the mathematical link to maximize the ELBO. \cite{Takahashi2018StudenttVA, melba:2022:008:akrami} assume their decoder distributions as Student-$t$ and asymmetric Laplace distributions, respectively, to mitigate the \textit{zero-variance problem} that the model training becomes unstable if the estimated variance of the decoder shrinks to zero in Gaussian VAE \cite{Lucas2019DontBT, NEURIPS2019_07211688, dai2018diagnosing}. 
\cite{Barron2017AGA} proposes a general distribution of the decoder, which allows improved robustness by optimizing the shape of the loss function during training. Recently, \cite{bredell2023explicitly} proposes a reconstruction loss that directly minimizes the \textit{blur error} of the VAE by modeling the covariance matrix of multivariate Gaussian decoder.

On the other hand, there exists a research direction that focuses on replacing the reconstruction loss without concern for losing the mathematical derivation of the lower bound. \cite{pmlr-v48-larsen16, Rosca2017VariationalAF, Munjal2019ImplicitDI} replace the reconstruction loss with an adversarial loss of the GAN framework. \cite{Hou2016DeepFC} introduces a feature-based loss that is calculated with a pre-trained convolutional neural network (CNN). Another approach by \cite{10.5555/3495724.3495897} adopts Watson's perceptual model, and \cite{Jiang2020FocalFL} directly optimizes the generative model in the frequency domain by a focal frequency reconstruction loss. Most of the above-mentioned methods aim to capture the properties of human perception by replacing the element-wise loss ($L_1$ or $L_2$-norm), which hinders the reconstruction of images \cite{pmlr-v48-larsen16}.

\textbf{Synthetic data generation.}
The GAN framework is widely adopted in the synthetic data generation task since it enables synthetic data generation in a nonparametric approach \cite{Choi2017GeneratingMD, Park2018DataSB, NEURIPS2019_254ed7d2, pmlr-v157-zhao21a}. \cite{NEURIPS2019_254ed7d2, pmlr-v157-zhao21a} assume that continuous columns of tabular datasets can be approximated by the Gaussian mixture distribution and model their decoder using Gaussian mixture distribution. Additionally, \cite{NEURIPS2019_254ed7d2, pmlr-v157-zhao21a} preprocess the continuous variables using the variational Gaussian mixture model \cite{Blei2016VariationalIA}, which is known as the mode-specific normalization technique. However, the preprocessing step requires additional computational resources and hyperparameter tuning of the number of modes. Other approaches by \cite{Park2018DataSB, pmlr-v157-zhao21a} regularize the discrepancy between the first and second-order statistics of the observed and synthetic dataset. \cite{Choi2017GeneratingMD} proposes the GAN-based synthesizer, which focuses on generating high-dimensional discrete variables with the assistance of the pre-trained AutoEncoder. 

\section{Proposal}
\label{sec:3}

Let $\bx \in \mbR^{p+q}$ be an observation consisting of continuous and discrete variables and $I = I_C \cup I_D = \{1, \cdots, (p+q)\}$ be an index set of the variables, where $I_C$ and $I_D$ correspond to index sets of $p$ continuous and $q$ discrete variables. $T_j$ denotes the number of levels for the discrete variables $\bx_j, j \in I_D$. We denote the ground-truth underlying distribution (probability density function, PDF) as $p(\bx)$ and the ground-truth CDF as $F(\bx)$. 

Let $\bz$ be a latent variable, where $\bz \in \mbR^d$ and $d < p+q$. The prior and posterior distribution of $\bz$ are assumed to be $p(\bz) = \mathcal{N}(\bz|\mathbf{0}, \mathbf{I})$ and $q(\bz|\bx;\phi) = \mathcal{N}\big(\bz | \mu(\bx;\phi), diag(\sigma^2(\bx;\phi))\big)$, respectively. Here, $\mathbf{I}$ is $d \times d$ identity matrix, $\mu:\mbR^{p+q} \mapsto \mbR^d$, $\sigma^2:\mbR^{p+q} \mapsto \mbR_+^d$ are neural networks parameterized with $\phi$, and $diag(a), a \in \mbR^d$ denotes a diagonal matrix with diagonal elements $a$. Moreover, we consider $\alpha \in [0, 1]$ as a random variable having density $p(\alpha)$.

\subsection{Distributional Learning}
\label{sec:3.1}

Our proposed model assumes that $p(\bx)$ is parametrized by an infinite mixture of asymmetric Laplace distribution (ALD) \cite{brando2019modelling}. The ALD is characterized by two parameters: $\alpha$, representing the asymmetry, and $\beta > 0$, representing the scale. By considering these parameters, along with the model parameter $\theta$, we can define the probability model of $\bx$ as follows:
\bean
p(\bx;\theta, \beta) = \int \int_0^1 p(\bx | \bz , \alpha; \theta, \beta) p(\bz, \alpha) d\alpha d\bz.
\eean

\begin{assumption} \label{assump0}
(1) $\{\bx_j\}_{j \in I}$ are conditionally independent given $\bz$. 
(2) The discrete random variables $\{\bx_j\}_{j \in I_D}$ are independent of $\alpha$.
(3) $\alpha$ and $\bz$ are independent.
\end{assumption}
By Assumption \ref{assump0}-(1), we model the dependency between $\bx_j$s solely through the latent variable $\bz$ \cite{Peters2017ElementsOC}. Assumption \ref{assump0}-(2) implies that $\alpha$ is related only to the continuous variables. Then, the decoder of our VAE model denoted as $p(\bx | \bz, \alpha; \theta, \beta)$ is specified by equation \eqref{eq:decoder}:
\bea \label{eq:decoder}
p(\bx|\bz,\alpha;\theta,\beta) &=& \prod_{j\in I_C} p(\bx_j|\bz,\alpha;\theta_j,\beta) \cdot \prod_{j \in I_D} p(\bx_j|\bz;\theta_j) \\
&=& \prod_{j \in I_C} \frac{\alpha (1 - \alpha)}{\beta} \exp \left( -\rho_{\alpha}\left( \frac{\bx_j - D_j(\alpha,\bz; \theta_j)}{\beta} \right) \right) \cdot \prod_{j \in I_D} \prod_{l=1}^{T_j} \pi_l(\bz; \theta_j)^{\mathbb{I} (\bx_j = l)}, \nonumber
\eea
where $\theta = (\theta_1, \cdots, \theta_{p+q})$, $\beta$ is a non-trainable constant, $\rho_{v}(u) = u(v - \mathbb{I}(u < 0))$ (check function), and $\mathbb{I}(\cdot)$ denotes the indicator function. $D_j(\cdot,\cdot; \theta_j): [0, 1] \times \mbR^d \mapsto \mbR$ is the location parameter of ALD, which is parameterized with $\theta_j$ \cite{melba:2022:008:akrami}. For discrete variables, $\pi(\cdot; \theta_j): \mbR^d \mapsto \Delta^{T_j-1}$ is a neural network parameterized with $\theta_j$, where $\Delta^{T_j-1}$ is the standard $(T_j-1)$-simplex for all $\bz \in \mbR^d$, and the subscript $l$ referes to the $l$th element of the output $\pi$. 

Assumption \ref{assump0}-(3)
leads our objective function, 
\bea \label{eq:final_obj}
\min_{\theta,\phi} && \mbE_{p(\bx)} \mbE_{q(\bz|\bx;\phi)} \left[ \sum_{j \in I_C}  \int_0^1 \rho_{\alpha}\Big( \bx_j - D_j(\alpha,\bz; \theta_j) \Big) d\alpha - \sum_{j \in I_D} \sum_{l=1}^{T_j} \mathbb{I}(\bx_j = l) \cdot \log \pi_l(\bz;\theta_j) \right] \nonumber\\
&+& \beta \cdot \mbE_{p(\bx)} [\mathcal{KL}(q(\bz|\bx;\phi) \| p(\bz))],
\eea
where constant terms are omitted. 
In order to achieve balanced learning of the two reconstruction losses in \eqref{eq:final_obj}, we have removed the weight $\beta$ associated with the second reconstruction loss. We refer to our model as `DistVAE.' The first term in \eqref{eq:final_obj} corresponds to the CRPS loss, which measures the accuracy of the proposed CDF approximation with respect to the ground-truth CDF of the underlying distribution \cite{Gneiting2007StrictlyPS, Matheson1976ScoringRF, gasthaus2019probabilistic}.

Interestingly, \eqref{eq:final_obj} is the limit of the negative ELBO derived from a finite mixture of ALD. We introduce $\alpha$, a discrete uniform random variable taking values on $\alpha_k = k/K$ for $k=1,\cdots,K$.
Then, the negative ELBO of $p(\bx;\theta, \beta)$ scaled with $\beta$ is written by
\bea \label{eq:negELBO}
&& \mbE_{q(\bz|\bx;\phi)} \left[ \sum_{j \in I_C} \frac{1}{K} \sum_{k=1}^{K} \rho_{\alpha_k}\Big( \bx_j - D_j(\alpha_k,\bz; \theta_j) \Big) \right] - \beta \frac{p}{K} \sum_{k=1}^{K} \log \alpha_k (1 - \alpha_k) + \beta p \log \beta \nonumber\\
&-& \beta \cdot \mbE_{q(\bz|\bx;\phi)} \left[\sum_{j \in I_D} \sum_{l=1}^{T_j} \mathbb{I}(\bx_j = l) \cdot \log \pi_l(\bz;\theta_j) \right] + \beta \cdot \mathcal{KL}(q(\bz|\bx;\phi) \| p(\bz))
\eea
(see Appendix \ref{app:1} for detailed derivation). 
The reconstruction loss, which corresponds to the first term of \eqref{eq:negELBO}, is a composite quantile loss for estimating the target quantiles $\alpha_k$s \cite{Yu2001BayesianQR, Koenker2017HandbookOQ, Moon2021LearningMQ, Wen2017AMQ, Cannon2018NoncrossingNR}. 
This entails adopting a Bayesian perspective for $\alpha$ as a prior (the Bayesian modeling for estimating multiple quantiles). Furthermore, throughout the derivation of the reconstruction loss, the role of $\alpha$ is pivotal in ensuring the representation of the reconstruction loss. To prevent the observation $\bx$ from influencing the distribution of $\alpha$, $\alpha$ is only assigned with a prior distribution, and the resulting reconstruction loss becomes the CRPS loss, a proper scoring rule.

However, for distributional learning of VAE, it is necessary to estimate conditional quantiles for an infinite number of quantile levels, denoted by $K \rightarrow \infty$ \cite{plataniotis2017gaussian}. The subsequent Theorem \ref{prop:convergence} establishes the convergence of the negative ELBO \eqref{eq:negELBO} to our objective function \eqref{eq:final_obj} as $K \rightarrow \infty$ \cite{brando2019modelling}.

\begin{theorem} \label{prop:convergence}
For all $j \in I_C$, suppose that $\int_0^1 \mbE_{p(\bx)} \mbE_{q(\bz|\bx;\phi)} \rho_{\alpha}\big(\bx_j - D_j(\alpha,\bz; \theta_j)\big) d\alpha < \infty$, and $\mbE_{p(\bx)} \mbE_{q(\bz|\bx;\phi)} [\rho_{\alpha}(\bx_j - D_j(\alpha,\bz; \theta_j))]$ is continuous over $\alpha \in (0, 1)$. Then, 
\bean
\lim_{K \rightarrow \infty} \mbE_{p(\bx)} \mbE_{q(\bz|\bx;\phi)} \left[ \frac{1}{K} \sum_{k=1}^{K} \rho_{\alpha_k}\Big( \bx_j - D_j(\alpha_k,\bz; \theta_j) \Big) \right] = \mbE_{p(\bx)} \mbE_{q(\bz|\bx;\phi)} \left[ \int_0^1 \rho_{\alpha}\Big( \bx_j - D_j(\alpha,\bz; \theta_j) \Big) d\alpha \right],
\eean
and $\lim_{K \rightarrow \infty} \frac{1}{K} \sum_{k=1}^{K} \log \alpha_k (1 - \alpha_k) = \int_0^1 \log \alpha (1-\alpha) d\alpha = -2$.
\end{theorem}

\subsection{Theoretical Results}
\label{sec:3.2}

In this section, we aim to provide theoretical insights into the ability of DistVAE, utilizing the objective function \eqref{eq:final_obj}, to recover the ground-truth distribution $p(\bx)$. To simplify the analysis without loss of generality, we consider the scenario where $\bx$ comprises only $p$ continuous random variables. 
Hence, we have $I = I_C = \{1, \cdots, p\}$, and $p(\bx)$ is defined over $\bx \in \mbR^p$ with $p(\bx) > 0$ for all $\bx \in \mbR^p$.
First, define a function $q(\bx|\bz;\phi)$ by
\bean
q(\bx|\bz;\phi) \coloneqq \frac{p(\bx)q(\bz|\bx;\phi)}{q(\bz;\phi)},
\eean
where $q(\bz;\phi) \coloneqq \int q(\bz|\bx;\phi) p(\bx) d\bx$ is the aggregated posterior \cite{pmlr-v84-tomczak18a}. Clearly, $q(\bx|\bz;\phi)$ is a PDF of $\bx$ for a given $\bz$. $q(\bx|\bz;\phi)$ is a conditional PDF of $\bx$ parametrized by $\phi$ and it is an approximated PDF of $\int_0^1 p(\bx|\bz, \alpha; \theta, \beta) d\alpha$. 
Since we assume that $\bx_j$s are conditionally independent in Assumption \ref{assump0}, $q(\bx|\bz;\phi) = \prod_{j=1}^p q_j(\bx_j|\bz;\phi)$ and the conditional CDF is written as 
\bea \label{eq:cond_cdf}
F(\bx|\bz;\phi) = \prod_{j=1}^p F_j(\bx_j|\bz;\phi), \mbox{\ where \ } F_j(\bx_j|\bz;\phi) \coloneqq \int_{-\infty}^{\bx_j} q_j(x|\bz;\phi) dx.
\eea
For notational simplicity, we let 
\bean
\theta^*(\phi) \in \arg\min_\theta \mbE_{p(\bx) q(\bz|\bx;\phi)} \sum_{j=1}^p \int_0^1 \rho_\alpha (\bx_j - D_j(\alpha,\bz;\theta_j)) d\alpha,
\eean
where $\theta^*(\phi) = (\theta_1^*(\phi),\cdots,\theta_p^*(\phi))$.

\begin{assumption} \label{assump1}
(1) Given an arbitrary $\phi$, $F_j(\cdot|\bz;\phi): \mbR \mapsto [0, 1]$ is absolutely continuous and strictly monotone increasing for all $j=1,\cdots,p$, and $\bz \in \mbR^d$. 
(2) Given an arbitrary $\theta$, $D_j(\cdot,\bz; \theta_j)$ is invertible and differentiable for all $j=1,\cdots,p$ and $\bz \in \mbR^d$.
(3) The aggregated posterior $q(\cdot;\phi)$ is absolutely continuous to the prior distribution of $\bz$.
\end{assumption}

\begin{theorem} \label{prop:main1}
Under Assumption \ref{assump1}, for an arbitrary $\phi$,
\bean
\mathcal{KL}\left( p(\bx) \Bigg\| \int \prod_{j=1}^p \frac{d}{d\bx_j} D_j^{-1}(\bx_j,\bz;\theta_j^*(\phi)) q(\bz;\phi) d\bz \right) = 0.
\eean
\end{theorem}
Theorem \ref{prop:main1} shows that DistVAE is capable of recovering the ground-truth distribution $p(\bx)$, indicating its ability to facilitate distributional learning rather than data reconstruction. Nevertheless, relying on the aggregated posterior distribution may lead to overfitting \cite{hoffman2016elbo, makhzani2015adversarial}, and sampling from the aggregated posterior can introduce computational challenges due to the absence of a straightforward closed-form representation for $q(\bz;\phi)$. To address these concerns, we propose an alternative approach that leverages the prior distribution $p(\bz)$ instead of $q(\bz;\phi)$, thereby enabling a computationally efficient synthetic generation process. This is substantiated by Theorem \ref{prop:main2}.

We define the estimated PDF $\hat{p}(\bx;\theta^*(\phi))$ and CDF $\hat{F}(\bx;\theta^*(\phi))$ as 
\bea \label{eq:estpdf}
\hat{p}(\bx;\theta^*(\phi)) &\coloneqq& \int \prod_{j=1}^p \frac{d}{d\bx_j} D_j^{-1}(\bx_j,\bz;\theta_j^*(\phi)) p(\bz) d\bz \\
\label{eq:estcdf}
\hat{F}(\bx; \theta^*(\phi)) &\coloneqq& \int \prod_{j=1}^p D_j^{-1}(\bx_j,\bz;\theta_j^*(\phi)) p(\bz) d\bz.
\eea

\begin{theorem} \label{prop:main2}
Suppose that $\phi$ is given such that $\mathcal{KL}(q(\bz;\phi) \| p(\bz)) < \epsilon$ for any $\epsilon > 0$. Then, under Assumption \ref{assump1}, 
\bean
\mathcal{KL}\left( p(\bx) \Bigg\| \int \prod_{j=1}^p \frac{d}{d\bx_j} D_j^{-1}(\bx_j,\bz;\theta_j^*(\phi)) p(\bz) d\bz \right) < \epsilon.
\eean
\end{theorem}

Theorem \ref{prop:main2} shows that even if we use the prior distribution $p(\bz)$ instead of the aggregated posterior $q(\bz;\phi)$, it is feasible to minimize the KL-divergence between the ground-truth PDF $p(\bx)$ and our estimated PDF $\hat{p}(\bx;\theta^*(\phi))$ of \eqref{eq:estpdf}. This is achievable because the KL-divergence term in \eqref{eq:final_obj} is the upper bound of $\mathcal{KL}(q(\bz;\phi) \| p(\bz))$ and is minimized during the training process. Since each conditional distribution of the estimated PDF depends on the same latent variable, it can be seen that the correlation structure between covariates is modeled implicitly.

\cite{higgins2017betavae, Lucas2019DontBT, kumar2020implicit} have highlighted the role of the KL-divergence weight parameter $\beta$ in controlling the precision of reconstruction. In our case, where the reconstruction loss is based on the CRPS loss, an increase in $\beta$ leads to a less accurate estimation of the ground-truth CDF. It implies that a large $\beta$ corresponds to lower-quality synthetic data, but it also enhances privacy level. Thus, $\beta$ introduces a trade-off between the quality of synthetic data and the risk of privacy leakage. The privacy level can be adjusted by manipulating $\beta$ \cite{Park2018DataSB}, as demonstrated in the experimental results presented in Section \ref{sec:4}.

\subsubsection{Synthetic Data Generation}
\label{sec:3.2.1}

By estimating the conditional quantile functions, we can transform the synthetic data generation process into inverse transform sampling. This conversion offers a notable advantage as it provides a straightforward and efficient approach to generating synthetic data. We denote a synthetic sample of $\bx_j$ as $\hat{x}_j$ for $j \in I$, and the synthetic data generation process can be summarized as follows:
\ben
    \item Sampling from the prior distribution: $z \sim p(\bz)$.
    \item Inverse transform sampling: For $j \in I_C$, $\hat{x}_j = D_j(u_j|z;\theta_j)$, where $u_j \sim U(0, 1)$.
    \item Gumbel-Max trick \cite{gumbel1954statistical}: For $j \in I_D$, $\hat{x}_j = \arg\max_{l=1,\cdots,T_j} \{\log \pi_l(z;\theta_j) + G_l\}$, where $G_l \sim Gumbel(0, 1)$, and $l=1,\cdots,T_j$.
\een

Note that both continuous and discrete variables share the same latent variable $z$. This shared latent variable allows for capturing the dependencies between variables. We numerically observe that the sampling, implemented using the Gumbel-Max trick, maintains the imbalanced ratio of the labels in the discrete variable.

\subsubsection{Parameterization of ALD}
\label{sec:3.2.2}

As introduced in \cite{gasthaus2019probabilistic}, for $j \in I_C$, we parameterize the function $D_j$, the location parameter of ALD, by a linear isotonic spline as follows:
\bea \label{eq:isotonic}
D_j(\alpha,\bz;\theta_j) = \gamma^{(j)}(\bz) + \sum_{m=0}^M b_m^{(j)} (\bz) (\alpha - d_m)_+ \mbox{\quad s.t. \quad} \sum_{m=0}^k b_m^{(j)}(\bz) \geq 0, k=1,\cdots,M,
\eea
where $\gamma^{(j)}(\bz) \in \mbR$, $b^{(j)}(\bz) = (b_0^{(j)}(\bz), \cdots, b_M^{(j)}(\bz)) \in \mbR^{M+1}$, $d = (d_0, \cdots, d_M) \in [0, 1]^{M+1}$, $0 = d_0 < \cdots < d_M = 1$, and $(u)_+ \coloneqq \max(0, u)$. $\theta_j$ is a neural network parameterized mapping such that $\theta_j: \mbR^d \mapsto \mbR \times \mbR^{M+1}$, which takes $\bz$ as input and outputs $\gamma^{(j)}(\bz)$ and $b^{(j)}(\bz)$. The constraint is introduced to ensure monotonicity. 
As demonstrated in \cite{gasthaus2019probabilistic}, the reconstruction loss can be computed in a closed form by utilizing the parameterization of \eqref{eq:isotonic} (refer to Appendix \ref{app:loss} for a detailed description of the loss function). This implies that our objective function \eqref{eq:final_obj} is computationally tractable. Note that the linear isotonic spline is not differentiable for finite points where the measure has no point mass.

\section{Experiments}
\label{sec:4}

\subsection{Overview}

\textbf{Dataset.} For evaluation, we consider following real tabular datasets: \texttt{covertype}, \texttt{credit}, \texttt{loan}, \texttt{adult}, \texttt{cabs}, and \texttt{kings} (see Appendix \ref{app:7} for detailed data descriptions). We treat the ordinal variables as continuous variables and discretize the estimated CDF (see Appendix \ref{app:5} for the discretization algorithm). Synthetic samples of ordinal variables are rounded to the first decimal place. 

\textbf{Compared models.} We compare DistVAE\footnote{We release the code at \url{https://github.com/an-seunghwan/DistVAE}.} with the state-of-the-art synthesizers; CTGAN \cite{NEURIPS2019_254ed7d2}, TVAE \cite{NEURIPS2019_254ed7d2}, and CTAB-GAN \cite{pmlr-v157-zhao21a}. All models have the same size of the latent dimension ($d=2$). 
The chosen latent space indeed limits the capacity of decoders for all models. However, we maintain a small and consistent number of parameters across all models during the experiment to isolate the performance differences in synthetic data generation to the methodologies of each synthesizer, specifically emphasizing the contribution of the decoder model's flexibility in estimating underlying distributions (see Table \ref{table:num_params} in Appendix \ref{app:8} for a comprehensive comparison of the model parameters).

\subsection{Evaluation Metrics}

To assess the quality of the synthetic data, we employ three types of assessment criteria: 1) machine learning utility, 2) statistical similarity, and 3) privacy preservability. Each criterion is evaluated using multiple metrics, and the performance of synthesizers is reported by averaged metrics over the real tabular datasets. The synthetic dataset is generated to have an equal number of samples as the real training dataset.

\textbf{Machine learning utility.} The machine learning utility (MLu) is measured by the predictive performance of the trained model over the synthetic data. We consider three popular machine learning algorithms: linear (logistic) regression, Random Forest \cite{breiman2001random}, and Gradient Boosting \cite{friedman2001greedy}. We measure the performance by utilizing the mean absolute relative error (MARE) for regression tasks \cite{Park2018DataSB} and the $F_1$ score for classification tasks \cite{NEURIPS2019_254ed7d2, pmlr-v157-zhao21a, wen2022causaltgan, Kamthe2021CopulaFF, Park2018DataSB, Choi2017GeneratingMD, Fang2022OvercomingCO}. 

\textbf{Statistical similarity.} 
The marginal distributional similarity between the real training and synthetic datasets is evaluated using two metrics: the Kolmogorov statistic and the 1-Wasserstein distance \cite{Fang2022OvercomingCO}. These metrics measure the distance between the empirical marginal CDFs \cite{Lehmann1998ElementsOL}. The joint distributional similarity is assessed by comparing the correlation matrices \cite{pmlr-v157-zhao21a}. To compute the correlation matrix and measure the $L_2$ distance between the correlation matrices of the real training and synthetic datasets, we employ the \texttt{dython} library \footnote{\url{http://shakedzy.xyz/dython/modules/nominal/\#associations}}. These enable a comprehensive evaluation of both marginal and joint distributional similarities between the real training and synthetic datasets.

\textbf{Privacy preservability.} 
The privacy-preserving capacity is measured using three metrics: the \textit{distance to closest record} (DCR) \cite{Park2018DataSB, pmlr-v157-zhao21a}, \textit{membership inference attack} \cite{Shokri2016MembershipIA, Choi2017GeneratingMD, Park2018DataSB}, and \textit{attribute disclosure} \cite{Choi2017GeneratingMD, Matwin2015ARO}. As in \cite{pmlr-v157-zhao21a}, the DCR is defined as the $5^{th}$ percentile of the $L_2$ distances between all real training samples and synthetic samples. Since the DCR is a $L_2$ distance-based metric, it is computed using only continuous variables. A higher DCR value indicates a more effective preservation of privacy, indicating a lack of overlap between the real training data and the synthetic samples. Conversely, an excessively large DCR score suggests a lower quality of the generated synthetic dataset. Therefore, the DCR metric provides insights into both the privacy-preserving capability and the quality of the synthetic dataset.

The membership inference attack evaluation follows the steps detailed in Appendix \ref{app:6}. The procedure is customized to be applied to a VAE-based synthesizer, such as DistVAE and TVAE. By transforming the problem into a binary classification task, we aim to identify the intricate relationship between the real training data and the synthetic samples. Higher binary classification scores indicate a higher vulnerability of the target synthesizer to membership inference attacks. 

Attribute disclosure refers to the situation where attackers can uncover additional covariates of a record by leveraging a subset of covariates they already possess, along with similar records from the synthetic dataset. To quantify the extent to which attackers can accurately identify these additional covariates, we employ classification metrics. Higher attribute disclosure metrics indicate an increased risk of privacy leakage, implying that attackers can precisely infer unknown variables. In terms of privacy concerns, attribute disclosure can be considered a more significant issue than membership inference attacks, as attackers are assumed to have access to only a subset of covariates for a given record \cite{Choi2017GeneratingMD}.

\subsection{Results}

\textbf{Machine learning utility.} 
We expect a high-quality synthesizer to generate synthetic data with comparable predictive performance to the real training dataset, denoted as the `Baseline' in Table \ref{tab:mlu}. The results in Table \ref{tab:mlu} demonstrate that DistVAE achieves a competitive MARE score and outperforms other methods in terms of the $F_1$ score. Furthermore, the performance of MLu improves as the value of $\beta$ decreases, indicating that the quality of the generated synthetic data is controlled by $\beta$. For a comprehensive overview of the MLu scores for all tabular datasets, please refer to Appendix \ref{app:9}.

\begin{table}[ht]
\begin{minipage}[t]{.57\textwidth}
\caption{Averaged MLu metrics (MARE, $F_1$). Mean and standard deviation values are obtained from 10 repeated experiments. $\uparrow$ denotes higher is better and $\downarrow$ denotes lower is better.}
  \centering
  \begin{tabular}{lrr}
    \toprule
    Model & MARE $\downarrow$ & $F_1$ $\uparrow$\\
    \midrule
Baseline & $0.150_{\pm 0.200}$ & $0.814_{\pm 0.101}$\\
CTGAN & $0.321_{\pm 0.271}$ & $0.672_{\pm 0.234}$\\
TVAE & $\textbf{0.225}_{\pm 0.215}$ & $0.594_{\pm 0.295}$\\
CTAB-GAN & $0.403_{\pm 0.392}$ & $0.702_{\pm 0.162}$\\
    \midrule
DistVAE($\beta=0.5$) & $0.349_{\pm 0.328}$ & $\textbf{0.769}_{\pm 0.128}$\\
DistVAE($\beta=1$) & $0.344_{\pm 0.316}$ & $\textbf{0.762}_{\pm 0.134}$\\
DistVAE($\beta=5$) & $0.392_{\pm 0.348}$ & $0.679_{\pm 0.190}$\\
    \bottomrule
  \end{tabular}
\label{tab:mlu}
\end{minipage}
\hfill
\begin{minipage}[t]{0.4\textwidth}
\caption{Averaged correlation structural similarity. `CorrDist' represents $L_2$ distance between the correlation matrices of synthetic and real training datasets. Mean and standard deviation values are obtained from 10 repeated experiments. Lower is better.}
  \centering
  \begin{tabular}{lr}
    \toprule
    Model & CorrDist \\
    \midrule
CTGAN & $2.180_{\pm 0.467}$\\
TVAE & $2.739_{\pm 0.796}$\\
CTAB-GAN & $2.575_{\pm 0.513}$\\
    \midrule
DistVAE($\beta=0.5$) & $\textbf{1.473}_{\pm 0.398}$\\
DistVAE($\beta=1$) & $1.730_{\pm 0.548}$\\
DistVAE($\beta=5$) & $3.113_{\pm 1.119}$\\
    \bottomrule
  \end{tabular}
\label{tab:corr}
\end{minipage}
\end{table}

\begin{figure*}[ht]
    \centering
    \includegraphics[width=\textwidth]{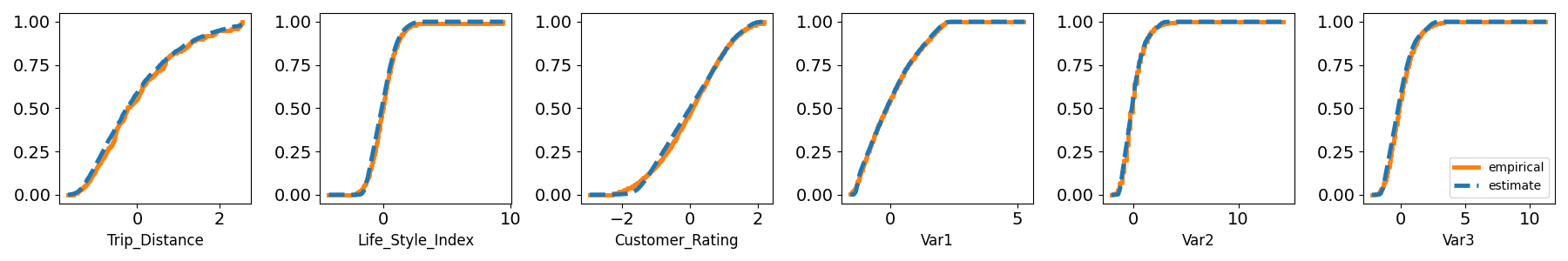}
    \caption{\texttt{cabs} dataset. Empirical (solid orange) and estimated (dashed blue) CDFs of continuous and ordinal variables (Monte Carlo approximated with 5000 samples). We standardize covariates and remove observations outside the 1\% and 99\% percentile range.}
    \label{fig:quantile}
\end{figure*}

\textbf{Statistical similarity.} 
The evaluation results for joint and marginal distributional similarities are presented in Table \ref{tab:corr} and \ref{tab:stat}. In Table \ref{tab:corr}, DistVAE achieves the lowest CorrDist score, indicating its ability to accurately preserve the correlation structure of the real training dataset in the generated synthetic dataset. Furthermore, DistVAE surpasses other methods in Table \ref{tab:stat} when it comes to marginal distributional similarity, suggesting that it successfully captures the underlying distribution of the observed dataset. Notably, reducing the value of $\beta$ leads to an enhancement in the quality of the synthetic dataset, as evidenced by improvements in the correlation structure and similarity of the marginal distributions. Figure \ref{fig:quantile} provides visualizations of the estimated CDFs \eqref{eq:estcdf} for each continuous (or ordinal) variable in the \texttt{cabs} dataset. For detailed statistical similarity scores and additional visualizations of estimated CDFs for all tabular datasets, please refer to Appendix \ref{app:9}.

\begin{table}[ht]
\caption{Averaged marginal distributional similarity. K-S denotes the Kolmogorov-Smirnov statistic, and 1-WD represents the 1-Wasserstein distance. Mean and standard deviation values are obtained from 10 repeated experiments. Lower is better.}
  \centering
  \begin{tabular}{lrrrr}
    \toprule
    & \multicolumn{2}{c}{Continuous} & \multicolumn{2}{c}{Discrete}\\
    \cmidrule(lr){2-3} \cmidrule(lr){4-5}
    Model & K-S & 1-WD & K-S & 1-WD \\
    \midrule
CTGAN & $0.133_{\pm 0.106}$ & $0.087_{\pm 0.025}$ & $0.168_{\pm 0.195}$ & $0.521_{\pm 0.532}$\\
TVAE & $0.196_{\pm 0.135}$ & $0.220_{\pm 0.099}$ & $0.385_{\pm 0.144}$ & $1.681_{\pm 1.668}$\\
CTAB-GAN & $0.157_{\pm 0.089}$ & $0.130_{\pm 0.037}$ & $0.106_{\pm 0.083}$ & $0.412_{\pm 0.378}$\\
    \midrule
DistVAE($\beta = 0.5$) & $0.090_{\pm 0.065}$ & $\textbf{0.075}_{\pm 0.026}$ & $0.030_{\pm 0.017}$ & $\textbf{0.118}_{\pm 0.100}$\\
DistVAE($\beta = 1$) & $\textbf{0.081}_{\pm 0.039}$ & $0.083_{\pm 0.019}$ & $\textbf{0.027}_{\pm 0.021}$ & $\textbf{0.116}_{\pm 0.110}$\\
DistVAE($\beta = 5$) & $0.092_{\pm 0.037}$ & $0.121_{\pm 0.058}$ & $0.059_{\pm 0.034}$ & $0.241_{\pm 0.163}$\\
    \bottomrule
  \end{tabular}
\label{tab:stat}
\end{table}

\textbf{Privacy preservability.} 
The privacy preservability performances of synthesizers, as measured by the DCR, are presented in Table \ref{tab:dcr}. DistVAE performs best in preserving privacy, with the highest DCR values compared to other methods. Notably, as the value of $\beta$ increases in DistVAE, the DCR between the real training and synthetic datasets (R\&S) also increases. This indicates that the risk of privacy leakage can be controlled by adjusting $\beta$, where higher values of $\beta$ correspond to a higher level of privacy protection. Moreover, DistVAE consistently achieves large DCR values for the synthetic dataset (S) across all $\beta$ values, indicating its ability to generate diverse synthetic samples. On the other hand, CTAB-GAN generates duplicated records in the synthetic dataset, resulting in relatively lower DCR scores for the synthetic dataset (S). For detailed DCR scores for all tabular datasets, please refer to Appendix \ref{app:9}.

\begin{table}[ht]
\caption{Privacy preservability: Averaged distance to closest record (DCR) between real training and synthetic datasets (R\&S), between the same real training datasets (R), and between the same synthetic datasets (S). Mean and standard deviation values are obtained from 10 repeated experiments. The DCR (R) score represents the baseline diversity of datasets. Higher is better.}
  \centering
  \begin{tabular}{lrrr}
    \toprule
    Model & R\&S & R & S \\
    \midrule
CTGAN & $0.426_{\pm 0.229}$ & $0.237_{\pm 0.153}$ & $0.356_{\pm 0.202}$\\
TVAE & $0.470_{\pm 0.181}$ & $0.237_{\pm 0.153}$ & $0.278_{\pm 0.195}$\\
CTAB-GAN & $0.508_{\pm 0.259}$ & $0.237_{\pm 0.153}$ & $0.039_{\pm 0.073}$\\
    \midrule
DistVAE($\beta=0.5$) & $0.444_{\pm 0.250}$ & $0.237_{\pm 0.153}$ & $0.463_{\pm 0.288}$\\
DistVAE($\beta=1$) & $0.463_{\pm 0.282}$ & $0.237_{\pm 0.153}$ & $0.479_{\pm 0.310}$\\ 
DistVAE($\beta=5$) & $\textbf{0.517}_{\pm 0.272}$ & $0.237_{\pm 0.153}$ & $\textbf{0.511}_{\pm 0.335}$\\
    \bottomrule
  \end{tabular}
\label{tab:dcr}
\end{table}

\begin{table}[ht]
\caption{Privacy preservability: Averaged membership inference attack performance. Mean and standard deviation values are obtained from 10 repeated experiments.}
  \centering
  \begin{tabular}{lrr}
    \toprule
    Model & Accuracy & AUC \\
    \midrule
TVAE & $0.495_{\pm 0.019}$ & $0.495_{\pm 0.019}$\\
DistVAE($\beta=0.5$) & $0.500_{\pm 0.003}$ & $0.500_{\pm 0.003}$\\
    \bottomrule
  \end{tabular}
\label{tab:memattack}
\end{table}

To evaluate the membership inference attack, we prepare one attack model per class. The attack testing records comprise an equal number of real training and test records, distinguished by the labels $in$ and $out$, respectively. Note that the test records are not employed in constructing the attack models. We employ gradient-boosting classifiers as the attack models, and for computational feasibility, we limit the number of attack models to one (i.e., $C=1$).

For the membership inference attack evaluation, we utilize accuracy and AUC (Area Under Curve) as the evaluation metrics. Since the target labels ($in/out$) are balanced, and the task is a binary classification problem, these metrics are appropriate. The results presented in Table \ref{tab:memattack} reveal that both DistVAE and TVAE achieve nearly identical accuracy and AUC scores of 0.5. This indicates that the attack models can \textit{not} distinguish between members of the real training and test datasets. Consequently, the membership inference attack is unsuccessful for both models. Therefore, DistVAE effectively generates synthetic datasets while ensuring privacy against membership inference attacks. A comprehensive assessment of membership inference attack performances for all tabular datasets can be found in Appendix \ref{app:9}.

\begin{table}[ht]
\caption{Privacy preservability: Averaged attribute disclosure performance with the $F_1$ score. Mean and standard deviation values are obtained from 10 repeated experiments. Lower is better.}
  \centering
  \begin{tabular}{lrrr}
    \toprule
    & \multicolumn{3}{c}{Number of neighbors ($k$)} \\
    \cmidrule(){2-4}
    Model & 1 & 10 & 100 \\
    \midrule
CTGAN & $0.262_{\pm 0.091}$ & $0.282_{\pm 0.087}$ & $0.275_{\pm 0.087}$\\
TVAE & $0.437_{\pm 0.162}$ & $0.438_{\pm 0.160}$ & $0.432_{\pm 0.162}$\\
CTAB-GAN & $\textbf{0.257}_{\pm 0.123}$ & $0.258_{\pm 0.114}$ & $0.261_{\pm 0.111}$\\
    \midrule
DistVAE($\beta = 0.5$) & $0.328_{\pm 0.088}$ & $0.328_{\pm 0.076}$ & $0.310_{\pm 0.072}$\\
DistVAE($\beta = 1$) & $0.307_{\pm 0.073}$ & $0.313_{\pm 0.068}$ & $0.297_{\pm 0.066}$\\
DistVAE($\beta = 5$) & $0.265_{\pm 0.105}$ & $\textbf{0.253}_{\pm 0.103}$ & $\textbf{0.232}_{\pm 0.101}$\\
    \bottomrule
  \end{tabular}
\label{tab:attrdis}
\end{table}

We present the attribute disclosure performance results in Table \ref{tab:attrdis}. For each value of $k$, we observe that as $\beta$ increases, the $F_1$ score of DistVAE decreases. Also, DistVAE achieves the smallest $F_1$ score when $k$ equals 10 and 100. Based on these results, we can conclude that DistVAE can generate synthetic datasets with a low risk of attribute disclosure, and the level of privacy preservation is controlled by $\beta$. Please refer to Appendix \ref{app:9} for a detailed evaluation of attribute disclosure performance for all tabular datasets.


\textbf{Quantile estimation.} To investigate the quantile estimation performance, we also evaluate DistVAE using the Vrate($\alpha$) metric \cite{RePEc:eee:intfor:v:28:y:2012:i:3:p:557-574}. The Vrate($\alpha$) is defined as $\frac{1}{|I_{test}|} \sum_{i \in I_{test}} I(x_i < \hat{Q}_\alpha)$, where $\alpha \in (0, 1)$, $I_{test}$ is the set of indices for the test dataset, $x_i$ is the $i$-th sample in the test dataset, and $\hat{Q}_\alpha$ is the empirical $\alpha$-quantile of the synthetic data. Since Vrate($\alpha$) indicates the proportion of compliance samples in the test dataset, the Vrate($\alpha$) score should be close to $\alpha$.

\begin{table}[ht]
\caption{Averaged Vrate($\alpha$) and $|\alpha - \mbox{Vrate}(\alpha)|$.}
  \centering
  \begin{tabular}{lrrrrr}
    \toprule
    $\alpha$ & 0.1 & 0.3 & 0.5 & 0.7 & 0.9 \\
    \midrule
Vrate($\alpha$) & 0.204 & 0.373 & 0.533 & 0.725 & 0.908 \\
$|\alpha - \mbox{Vrate}(\alpha)|$ & 0.104 & 0.083 & 0.04 & 0.032 & 0.008 \\
    \bottomrule
  \end{tabular}
\label{tab:alpharate}
\end{table}

The Vrate($\alpha$) evaluation results are presented in Table \ref{tab:alpharate}. Table \ref{tab:alpharate} shows that the ratio of violated test samples ($|\alpha - \mbox{Vrate}(\alpha)|$) decreases as $\alpha$ increases, indicating a better performance for estimating larger quantiles. However, the ratio of violated test samples is relatively large for smaller $\alpha$ values, which may be due to extremely skewed continuous variables, such as \texttt{capital-gain} and \texttt{capital-loss} from \texttt{adult} dataset, that make the quantile estimation unstable.

\section{Conclusion and Limitations}
\label{sec:5}

This paper introduces a novel distributional learning method of VAE, which aims to effectively capture the underlying distribution of the observed dataset using a nonparametric approach. Our proposed method involves directly estimating CDFs using the CRPS loss while maintaining the mathematical derivation of the lower bound.

In our study, we confirm that the proposed decoder enhances the performance of generative models for tabular data. However, this conclusion relies on the assumption of conditional independence among the observed variables given the latent variable. When the dimension of the latent variable is small, this assumption is prone to violation. Therefore, in cases where the size of the latent space is limited, the proposed nonparametric fitting of the decoder might not accurately represent the underlying distribution. Particularly in the image domain, where adjacent pixel values exhibit significant dependence, it remains uncertain whether our proposed model would lead to notable improvements in image data generation with a low-dimensional latent space.

Nevertheless, classical image datasets, such as CIFAR-10 \cite{Krizhevsky2009LearningML}, often exhibit pixel value distributions that deviate considerably from Gaussian, with the frequencies of edge values (0 and 255) dominating more than other pixel values \cite{darlow2018cinic, salimans2017pixelcnn}. Other image datasets, as presented in \cite{Parmar2022HardwareEfficientSB}, demonstrate multi-modality in pixel value distributions. These experimental findings suggest that leveraging the capacity of distributional learning could be advantageous in approximating the ground-truth distribution of image data when the latent variable effectively captures conditional independence among the image pixels. Consequently, we expect that compromising biases arising from violating conditional independence and marginally misspecified distributions may further enhance our results, and we leave it for future research.

On the other hand, we consider two approaches to enhance the performance of quantile estimation. Firstly, we plan to extend the parameterization of conditional quantile functions to a more flexible monotonic regression model. Secondly, we intend to incorporate the Uniform Pessimistic Risk (UPR) \cite{Hong2023UniformPR} into the VAE framework to handle lower quantile levels better. Furthermore, we are exploring the expansion of DistVAE into a time-series distributional forecasting model by adopting the conditional VAE framework \cite{Sohn2015LearningSO}. This extension will enable the application of our method to time-series data, opening new avenues for distributional forecasting.

\begin{ack}
This work was supported by the National Research Foundation of Korea (NRF) grant funded by the Korea government (MSIT) (No. NRF-2022R1A4A3033874 and No. NRF-2022R1F1A1074758). This work was also supported by Korea Environmental Industry \& Technology Institute (KEITI) through `Core Technology Development Project for Environmental Diseases Prevention and Management', funded by Korea Ministry of Environment (MOE) (2021003310005). The authors acknowledge the Urban Big data and AI Institute of the University of Seoul supercomputing resources (\url{http://ubai.uos.ac.kr}) made available for conducting the research reported in this paper.
\end{ack}

\bibliographystyle{plain}
\bibliography{ref}


\newpage
\appendix
\onecolumn
\section{Appendix}

\subsection{Derivation of ELBO}
\label{app:1}

\bean
&& \log p(\bx; \theta, \beta) \\
&=& \log \sum_{k=1}^{K} p(\alpha_k) \int p(\bx|\bz, \alpha_k;\theta,\beta) p(\bz) d\bz \\
&=& \log \sum_{k=1}^{K} p(\alpha_k) \int p(\bx|\bz, \alpha_k;\theta,\beta) \frac{p(\bz)}{q(\bz|\bx;\phi)} q(\bz|\bx;\phi) d\bz \\
&\geq& \sum_{k=1}^{K} p(\alpha_k) \int q(\bz|\bx;\phi) \log \Big( p(\bx|\bz, \alpha_k;\theta,\beta) \frac{p(\bz)}{q(\bz|\bx;\phi)} \Big) d\bz \\
&=& \frac{1}{K} \sum_{k=1}^{K} \mbE_{q(\bz|\bx;\phi)} [\log p(\bx|\bz, \alpha_k;\theta,\beta)] - \mathcal{KL}(q(\bz|\bx;\phi) \| p(\bz)) \\
&=& \frac{1}{K} \sum_{k=1}^{K} \mbE_{q(\bz|\bx;\phi)} \left[\sum_{j \in I_C} \log p(\bx_j|\bz, \alpha_k;\theta_j,\beta) + \sum_{j \in I_D} \log p(\bx_j|\bz;\theta_j,\beta)\right] - \mathcal{KL}(q(\bz|\bx;\phi) \| p(\bz)) \\
&=& \mbE_{q(\bz|\bx;\phi)} \left[ \sum_{j \in I_C} \frac{1}{K} \sum_{k=1}^{K} \log \frac{\alpha_k (1 - \alpha_k)}{\beta} -\rho_{\alpha_k}\left( \frac{\bx_j - D_j(\alpha_k|\bz, \theta_j)}{\beta} \right) \right] \\
&& + \mbE_{q(\bz|\bx;\phi)} \left[\sum_{j \in I_D} \sum_{l=1}^{T_j} I(\bx_j = l) \cdot \log \pi_l(\bz;\theta_j) \right] - \mathcal{KL}(q(\bz|\bx;\phi) \| p(\bz)) \\
&=& \mbE_{q(\bz|\bx;\phi)} \left[ - \frac{1}{\beta} \cdot \sum_{j \in I_C} \frac{1}{K} \sum_{k=1}^{K} \rho_{\alpha_k}\Big( \bx_j - D_j(\alpha_k|\bz, \theta_j) \Big) \right] + \frac{p}{K} \sum_{k=1}^{K} \log \alpha_k (1 - \alpha_k) - p \cdot \log \beta \\
&& + \mbE_{q(\bz|\bx;\phi)} \left[\sum_{j \in I_D} \sum_{l=1}^{T_j} I(\bx_j = l) \cdot \log \pi_l(\bz;\theta_j) \right] - \mathcal{KL}(q(\bz|\bx;\phi) \| p(\bz)) \\
&=& - \frac{1}{\beta} \Bigg( \mbE_{q(\bz|\bx;\phi)} \left[ \sum_{j \in I_C} \frac{1}{K} \sum_{k=1}^{K} \rho_{\alpha_k}\Big( \bx_j - D_j(\alpha_k|\bz, \theta_j) \Big) \right] - \beta \frac{p}{K} \sum_{k=1}^{K} \log \alpha_k (1 - \alpha_k) + \beta p \cdot \log \beta \\
&& - \beta \cdot \mbE_{q(\bz|\bx;\phi)} \left[\sum_{j \in I_D} \sum_{l=1}^{T_j} I(\bx_j = l) \cdot \log \pi_l(\bz;\theta_j) \right] + \beta \cdot \mathcal{KL}(q(\bz|\bx;\phi) \| p(\bz)) \Bigg),
\eean
by Jensen's inequality.


\clearpage
\subsection{Proof of Theorem \ref{prop:convergence}}
\label{app:2}


\begin{proof}
\bean
\mbE_{p(\bx)} \mbE_{q(\bz|\bx;\phi)} \left[ \frac{1}{K} \sum_{k=1}^{K} \rho_{\alpha_k}\Big( \bx_j - D_j(\alpha_k|\bz, \theta_j) \Big) \right] 
&=& \frac{1}{K} \sum_{k=1}^{K} h(\alpha_k) \\
&=& \sum_{k=1}^{K} h(\alpha_k) \cdot (\alpha_k - \alpha_{k-1}),
\eean
where $\alpha_0 \coloneqq 0$ and $\mbE_{p(\bx)} \mbE_{q(\bz|\bx;\phi)} \left[ \rho_{\alpha_k}\Big( \bx_j - D_j(\alpha_k|\bz, \theta_j) \Big) \right]$ is denoted as $h(\alpha_k)$.

Since $\alpha_k \in [\alpha_{k-1}, \alpha_k]$ and $h(\cdot): [0, 1] \mapsto \mbR$ is a continuous function, for $j \in I_C$,
\bean
\lim_{K \rightarrow \infty} \mbE_{p(\bx)} \mbE_{q(\bz|\bx;\phi)} \left[ \frac{1}{K} \sum_{k=1}^{K} \rho_{\alpha_k}\Big( \bx_j - D_j(\alpha_k|\bz, \theta_j) \Big) \right] &=& \lim_{K \rightarrow \infty} \sum_{k=1}^{K} h(\alpha_k) \cdot (\alpha_k - \alpha_{k-1}) \\
&=& \int_0^1 h(\alpha) d\alpha \\
&=& \int_0^1 \mbE_{p(\bx)} \mbE_{q(\bz|\bx;\phi)} \left[ \rho_{\alpha}\Big( \bx_j - D_j(\alpha|\bz, \theta_j) \Big) \right] d\alpha \\
&=& \mbE_{p(\bx)} \mbE_{q(\bz|\bx;\phi)} \left[ \int_0^1 \rho_{\alpha}\Big( \bx_j - D_j(\alpha|\bz, \theta_j) \Big) d\alpha \right],
\eean
by the definition of the Riemann integral and the Fubini-Tonelli theorem. The proof is complete.
\end{proof}

\subsection{Proof of Theorem \ref{prop:main1}}
\label{app:3}


\begin{proof}
\bean
\min_\theta \mbE_{p(\bx) q(\bz|\bx;\phi)} \sum_{j=1}^p \int_0^1 \rho_\alpha (\bx_j - D_j(\alpha,\bz;\theta_j)) d\alpha &=& \mbE_{q(\bz;\phi) q(\bx|\bz;\phi)} \sum_{j=1}^p \int_0^1 \rho_\alpha (\bx_j - D_j(\alpha,\bz;\theta_j)) d\alpha \\
&=& \sum_{j=1}^p \mbE_{q(\bz;\phi)} \mbE_{q(\bx_j|\bz;\phi)} \int_0^1 \rho_\alpha (\bx_j - D_j(\alpha,\bz;\theta_j)) d\alpha.
\eean
So, for all $j=1,\cdots,p$, 
\bean
\theta_j^*(\phi) \in \arg\min_{\theta_j} \mbE_{q(\bz;\phi)} \mbE_{q(\bx_j|\bz;\phi)} \int_0^1 \rho_\alpha (\bx_j - D_j(\alpha,\bz;\theta_j)) d\alpha,
\eean
and it is proper scoring rule relative to $F_j(\cdot|\bz;\phi)$ for all $\bz \in \mbR^d$. It implies that $D_j^{-1}(\bx_j,\bz;\theta_j^*(\phi)) = F_j(\bx_j|\bz;\phi)$, and $\frac{d}{d\bx_j} D_j^{-1}(\bx_j,\bz;\theta_j^*(\phi)) = q_j(\bx_j|\bz;\phi)$, for all $\bx_j \in \mbR$, by Assumption \ref{assump1}. 

It follows that
\bean
\int \prod_{j=1}^p \frac{d}{d\bx_j} D_j^{-1}(\bx_j,\bz;\theta_j^*(\phi)) q(\bz;\phi) d\bz &=& \int \prod_{j=1}^p q_j(\bx_j|\bz;\phi) q(\bz;\phi) d\bz \\
&=& \int q(\bx|\bz;\phi) q(\bz;\phi) d\bz \\
&=& \int \frac{p(\bx) q(\bz|\bx;\phi)}{q(\bz;\phi)} q(\bz;\phi) d\bz \\
&=& \int p(\bx) q(\bz|\bx;\phi) d\bz \\
&=& p(\bx).
\eean
The proof is complete.
\end{proof}

\subsection{Proof of Theorem \ref{prop:main2}}
\label{app:4}


\begin{proof}
\bean
\min_\theta \mbE_{p(\bx) q(\bz|\bx;\phi)} \sum_{j=1}^p \int_0^1 \rho_\alpha (\bx_j - D_j(\alpha,\bz;\theta_j)) d\alpha &=& \mbE_{q(\bz;\phi) q(\bx|\bz;\phi)} \sum_{j=1}^p \int_0^1 \rho_\alpha (\bx_j - D_j(\alpha,\bz;\theta_j)) d\alpha \\
&=& \sum_{j=1}^p \mbE_{q(\bz;\phi)} \mbE_{q(\bx_j|\bz;\phi)} \int_0^1 \rho_\alpha (\bx_j - D_j(\alpha,\bz;\theta_j)) d\alpha.
\eean
So, for all $j=1,\cdots,p$, 
\bean
\theta_j^*(\phi) \in \arg\min_{\theta_j} \mbE_{q(\bz;\phi)} \mbE_{q(\bx_j|\bz;\phi)} \int_0^1 \rho_\alpha (\bx_j - D_j(\alpha,\bz;\theta_j)) d\alpha,
\eean
and it is proper scoring rule relative to $F_j(\cdot|\bz;\phi)$ for all $\bz \in \mbR^d$. It implies that $D_j^{-1}(\bx_j,\bz;\theta_j^*(\phi)) = F_j(\bx_j|\bz;\phi)$, and $\frac{d}{d\bx_j} D_j^{-1}(\bx_j,\bz;\theta_j^*(\phi)) = q_j(\bx_j|\bz;\phi)$, for all $\bx_j \in \mbR$, by Assumption \ref{assump1}. 

It follows that 
\bean
\mathcal{KL}\left( p(\bx) \Bigg\| \int \prod_{j=1}^p \frac{d}{d\bx_j} D_j^{-1}(\bx_j,\bz;\theta_j^*(\phi)) p(\bz) d\bz \right) &=& \mathcal{KL}\left( p(\bx) \Bigg\| \int \prod_{j=1}^p q_j(\bx_j|\bz;\phi) p(\bz) d\bz \right) \\
&=& \mathcal{KL}\left( p(\bx) \Bigg\| \int q(\bx|\bz;\phi) p(\bz) d\bz \right) \\
&=& \mathcal{KL}\left( p(\bx) \Bigg\| \int \frac{p(\bx)q(\bz|\bx;\phi)}{q(\bz;\phi)} p(\bz) d\bz \right) \\
&=& \mathcal{KL}\left( p(\bx) \Bigg\| p(\bx) \int \frac{p(\bz) q(\bz|\bx;\phi)}{q(\bz;\phi)} d\bz \right) \\
&=& \mbE_{p(\bx)} \left[ - \log \int \frac{p(\bz) q(\bz|\bx;\phi)}{q(\bz;\phi)} d\bz \right] \\
&\leq& \mbE_{p(\bx)} \left[ \int q(\bz|\bx;\phi) \log \frac{q(\bz;\phi)}{p(\bz)} d\bz \right] \\
&=& \iint p(\bx) q(\bz|\bx;\phi) \log \frac{q(\bz;\phi)}{p(\bz)} d\bz d\bx \\
&=& \iint p(\bx) q(\bz|\bx;\phi) \log \frac{q(\bz;\phi)}{p(\bz)} d\bx d\bz \\
&=& \int q(\bz;\phi) \log \frac{q(\bz;\phi)}{p(\bz)} d\bz \\
&=& \mathcal{KL}(q(\bz;\phi) \| p(\bz)) \\
&<& \epsilon,
\eean
by Jensen's inequality, Fubini-Tonelli theorem, and the assumptions. The proof is complete.
\end{proof}

\subsection{Closed Form Loss}
\label{app:loss}

\bean
2 \cdot \int_0^1 \rho_{\alpha}\Big( \bx_j - D_j(\alpha,\bz, \theta_j) \Big) d\alpha
&=& (2 \Tilde{\alpha}_j - 1)\bx_j + (1 - 2\Tilde{\alpha}_j) \gamma^{(j)}(\bz) \\
&+& \sum_{m=1}^{M} b_m^{(j)}(\bz) \Bigg( \frac{1 - d_m^3}{3} - d_m 
- \max(\Tilde{\alpha}_j, d_m)^2 + 2 \max(\Tilde{\alpha}_j, d_m)d_m \Bigg),
\eean
where $D_j(\Tilde{\alpha}_j,\bz;\theta_j) = \bx_j$, $\Tilde{\alpha}_j = \frac{\bx_j - \gamma^{(j)}(\bz) + \sum_{m=0}^{m_0} b_m^{(j)}(\bz) d_m}{\sum_{m=0}^{m_0} b_m^{(j)}(\bz)}$, and $D_j(d_{m_0},\bz;\theta_j) \leq \bx_j \leq D_j(d_{m_0+1},\bz;\theta_j)$. 


\clearpage
\subsection{Discretization of Estimated CDF}
\label{app:5}

\begin{figure}[ht]
    \centering
    \includegraphics[width=0.7\linewidth]{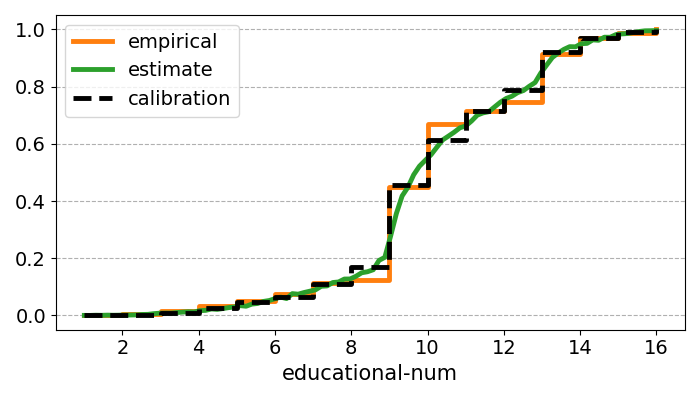}
    \caption{Discretized CDF for ordinal variable \texttt{educational-num} of \texttt{adult} dataset. `estimate' indicates $\hat{F}(\cdot;\theta)$, `calibration' indicates $\hat{F}^*(\cdot;\theta)$, and `empirical' indicates the empirical CDF of the observed dataset.}
    \label{fig:calibration}
\end{figure}

To ensure appropriate discretization of the estimated CDF for ordinal variables, we propose a post-ad-hoc discretization step \cite{salimans2017pixelcnn}. We focus on the case where $p=1$ and $q=0$ for brevity. We denote the set of observed possible values for the ordinal variable as ${x^{(1)}, x^{(2)}, \cdots, x^{(m)}}$. The discretization algorithm for the estimated CDF is presented in Algorithm \ref{alg:cal}, and we provide an example of the discretization algorithm's outcome in Figure \ref{fig:calibration}. 


\begin{algorithm}[ht]
\caption{Discretization of Estimated CDF} \label{alg:cal}
\begin{algorithmic}
\State \textbf{Input} $\{x^{(1)}, x^{(2)}, \cdots, x^{(m)}\}$, Estimated CDF $\hat{F}(\cdot;\theta)$
\State \textbf{Output} Discretized CDF $\hat{F}^*(\cdot;\theta)$
\State (1) Compute $\hat{F}(x^{(i)}-0.5;\theta)$ and $\hat{F}(x^{(i)}+0.5;\theta)$ for $i=1,\cdots,m$.
\State (2) Discretization: For $i=1,\cdots,m$,
\bean
\hat{F}^*(x^{(i)};\theta) &\coloneqq& \hat{F}^*(x^{(i-1)};\theta) \\
&+& \hat{F}(x^{(i)}+0.5;\theta) - \hat{F}(x^{(i)}-0.5;\theta),
\eean
where $\hat{F}^*(x^{(0)};\theta) \coloneqq 0$.
\State (3) Ensure monotonicity: For $i=1,\cdots,m-1$, if $\hat{F}^*(x^{(i)};\theta) > \hat{F}^*(x^{(i+1)};\theta)$,
\bean
\hat{F}^*(x^{(i+1)};\theta) \coloneqq \hat{F}^*(x^{(i)};\theta).
\eean
\end{algorithmic}
\end{algorithm}

\clearpage
\subsection{Membership Inference Attack}
\label{app:6}

\cite{Park2018DataSB} propose the customized \textit{membership inference attack} method of \cite{Shokri2016MembershipIA} to attack the GAN-based synthesizer. Similarly, we propose the customized membership inference attack method of \cite{Shokri2016MembershipIA} to attack the VAE-based synthesizer. 

\begin{assumption}[\cite{Shokri2016MembershipIA}] \label{assump:attack}
In the membership inference attack, the attacker attacks the target model under the following assumptions:
\bed
    \item[(A1)] The attacker is only allowed for black-box access, where the attacker can only supply inputs to the model and receive the model's output(s).
    \item[(A2)] The attacker can obtain as many outputs as they want from a target model to attack.
    \item[(A3)] The real and synthetic datasets should not have common records.
    \item[(A4)] The attacker knows the algorithm and architecture of the target model.
\eed
\end{assumption}

Denote $D_{train}^*$ and $D_{test}^*$ as the real training and test datasets. Under Assumption \ref{assump:attack}, the overall steps of the membership inference attack are outlined below:
\ben
    \item Generate shadow training and test datasets $D_{train}^{(i)}, D_{test}^{(i)}, i=1,\cdots,C$ from $M^*$ by (A1), where $M^*$ is the model attacker wants to attack. By (A2), the attacker is allowed to obtain shadow datasets such as $|D_{train}^{(i)}| = |D_{train}^*|$ and $D_{train}^{(i)} \cap D_{test}^{(j)} = \varnothing$, for $i=1,\cdots,C$. Under (A3), $D_{train}^{(i)} \cap D_{train}^* = \varnothing$, for $i=1,\cdots,C$.
    \item Train shadow models $M_1, \cdots, M_C$ under (A4), i.e., each shadow model is trained similarly to the target model $M^*$.
    \item For $i=1,\cdots,C$,
    \ben
        \item Obtain representation vectors $\bz$ from the encoder of $M_i$ with the input of $D_{train}^{(i)}$. Then, attacking training records are $(\by, \bz, in)$.
        \item Obtain representation vectors $\bz$ from the encoder of $M_i$ with the input of $D_{test}^{(i)}$. Then, attacking training records are $(\by, \bz, out)$.
    \een
    where $\by$ is the labels of shadow dataset records. And we assume that $\by$ consists of the MLu classification target.
    \item Merge all attack training records, $(\by, \bz, in/out)$.
    \item For each class of $\by$, train \textit{attack model} which is a binary classification model which classifies $in/out$ based on the representation vectors $\bz$.
    \item Now it is ready to attack.
\een
Note that we use the representation vector of VAE instead of the output of the GAN discriminator \cite{Park2018DataSB}. 

\clearpage
\subsection{Dataset Descriptions}
\label{app:7}

\textbf{Websites}
\bed
    \item \texttt{covertype}: \url{https://www.kaggle.com/datasets/uciml/forest-cover-type-dataset}
    \item \texttt{credit}: \url{https://www.kaggle.com/c/home-credit-default-risk}
    \item \texttt{loan}: \url{https://www.kaggle.com/datasets/teertha/personal-loan-modeling}
    \item \texttt{adult}: \url{https://www.kaggle.com/datasets/uciml/adult-census-income}
    \item \texttt{cabs}: \url{https://www.kaggle.com/datasets/arashnic/taxi-pricing-with-mobility-analytics?select=test.csv}
    \item \texttt{kings}: \url{https://www.kaggle.com/datasets/harlfoxem/housesalesprediction}
\eed

\begin{table*}[ht]
\caption{Description of datasets. \#C represents the number of continuous and ordinal variables. \#D denotes the number of discrete variables.}
  \centering
  \begin{tabular}{lrrrrr}
    \toprule
    Dataset & Train/Test Split & Regression Target & Classification Target & \#C & \#D \\
    \midrule
\texttt{covertype} & 45k/5k & \texttt{Elevation} & \texttt{Cover\_Type} & 10 & 1 \\
\texttt{credit} & 45k/5k & \texttt{AMT\_CREDIT} & \texttt{TARGET} & 10 & 9 \\
\texttt{loan} & 4k/1k & \texttt{Age} & \texttt{Personal Loan} & 5 & 6 \\
\texttt{adult} & 40k/5k & \texttt{age} & \texttt{income} & 5 & 9 \\
\texttt{cabs} & 40k/1k & \texttt{Trip\_Distance} & \texttt{Surge\_Pricing\_Type} & 6 & 7 \\
\texttt{kings} & 20k/1k & \texttt{long} & \texttt{condition} & 11 & 7 \\
    \bottomrule
  \end{tabular}
\label{tab:data_description}
\end{table*}

\clearpage
\subsection{Experimental Settings}
\label{app:8}

We run all experiments using Geforce RTX 3090 GPU, and our experimental codes are all available with \texttt{pytorch}. 

\begin{table*}[ht]
\caption{Hyper-parameter settings for tabular dataset experiments.}
\centering
  \begin{tabular}{lrrrrrr}
    \toprule
    Model & epochs & batch size & learning rate & $\beta$ (or decoder std range) & $d$ & $M$ \\
    \midrule
    CTGAN & 300 & 500 & 0.0002 & - & 2 & - \\
    TVAE & 200 & 256 & 0.005 & [0.1, 1] & 2 & -\\
    CTAB-GAN & 150 & 500 & 0.0002 & - & 2 & - \\
    \midrule
    DistVAE & 100 & 256 & 0.001 & 0.5 & 2 & 10 \\
    \bottomrule
  \end{tabular}
\end{table*}

\begin{table*}[ht]
\caption{The number of model parameters for tabular dataset experiments.}
\centering
  \begin{tabular}{lrrrrrr}
    \toprule
    Model & \texttt{covtype} & \texttt{credit} & \texttt{adult} & \texttt{loan} & \texttt{cabs} & \texttt{kings} \\
    \midrule
    CTGAN & 20k & 32k & 52k & 13k & 30k & 51k \\
    TVAE & 10k & 12k & 13k & 6k & 10k & 10k \\
    CTAB-GAN & 12k & 13k & 14k & 5k & 12k & 15k \\
    \midrule
    DistVAE & 10k & 12k & 13k & 6k & 10k & 16k \\
    \bottomrule
  \end{tabular}
\label{table:num_params}
\end{table*}

\begin{table}[ht]
\caption{Classifier and regressor used to evaluate synthetic data quality. The names of all parameters used in the description are consistent with those defined in corresponding packages.}
  \centering
  \resizebox{\columnwidth}{!}{
  \begin{tabular}{ccc}
    \toprule
    Tasks & Model & Description \\
    \midrule
    \multirow{5}{*}{\texttt{Regression}}
    & Linear Regression & \makecell{Package: \texttt{statsmodels.api.sm.OLS}, \\ setting: without intercept, defaulted values} \\
    & Random Forest & \makecell{Package: \texttt{sklearn.ensemble.RandomForestRegressor}, \\ setting: random\_state=0, defaulted values} \\
    & Gradient Boosting & \makecell{Package: \texttt{sklearn.ensemble.GradientBoostingRegressor}, \\ setting: random\_state=0, defaulted values} \\
    \midrule
    \multirow{5}{*}{\texttt{Classification}}
    & Logistic Regression & \makecell{Package: \texttt{sklearn.linear\_model.LogisticRegression}, \\ setting: multi\_class='ovr', fit\_intercept=False, defaulted values} \\
    & Random Forest & \makecell{Package: \texttt{sklearn.ensemble.RandomForestClassifier}, \\ setting: random\_state=0, defaulted values} \\
    & Gradient Boosting & \makecell{Package: \texttt{sklearn.ensemble.GradientBoostingClassifier}, \\ setting: random\_state=0, defaulted values} \\
    \bottomrule
  \end{tabular}}
\label{table:mlu_setting}
\end{table}

\clearpage
\subsection{Detailed Experimental Results}
\label{app:9}
\begin{figure*}[ht]
    \centering
    \subfigure[\texttt{covtype}]{
    \includegraphics[width=\textwidth]{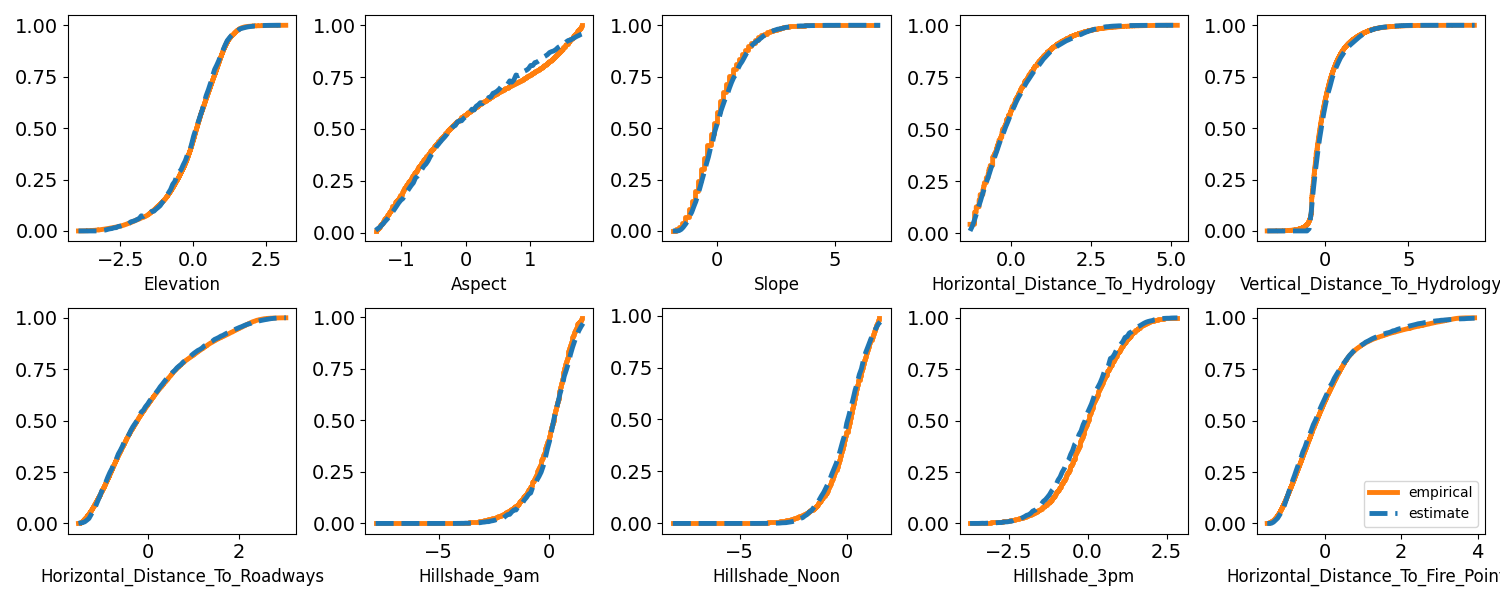}}
    \subfigure[\texttt{credit}]{
    \includegraphics[width=\textwidth]{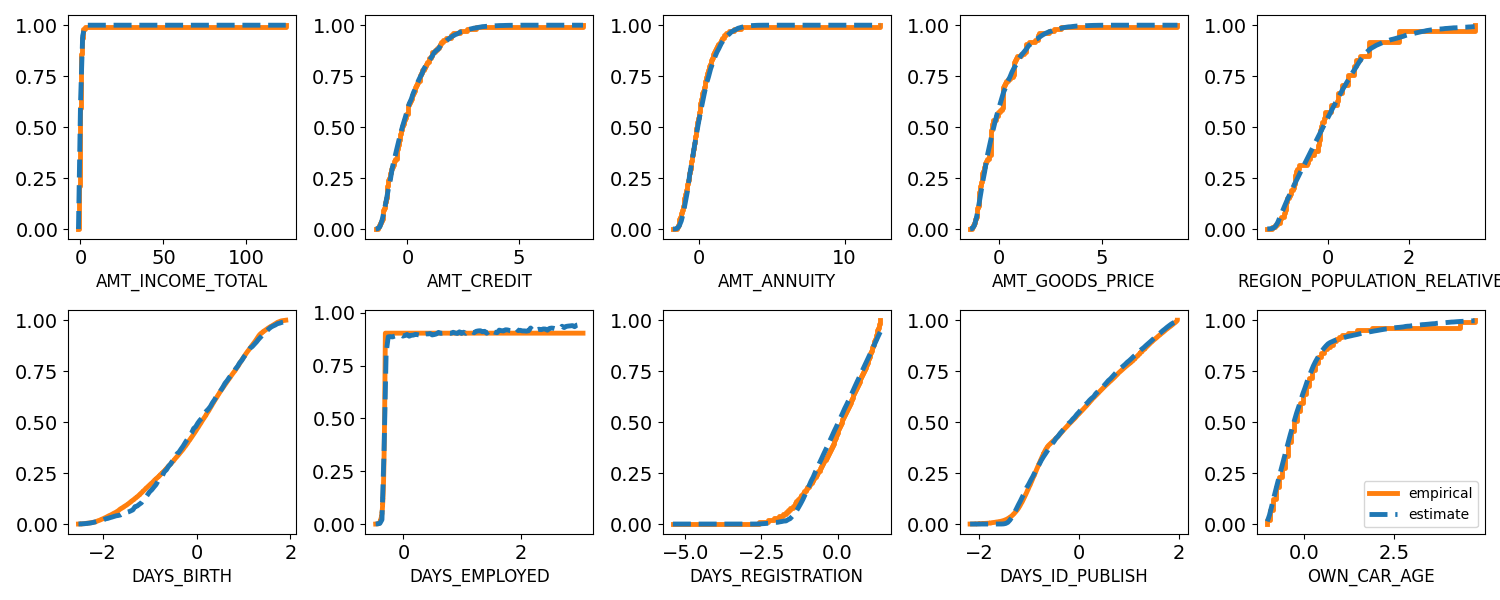}}
    \subfigure[\texttt{loan}]{
    \includegraphics[width=\textwidth]{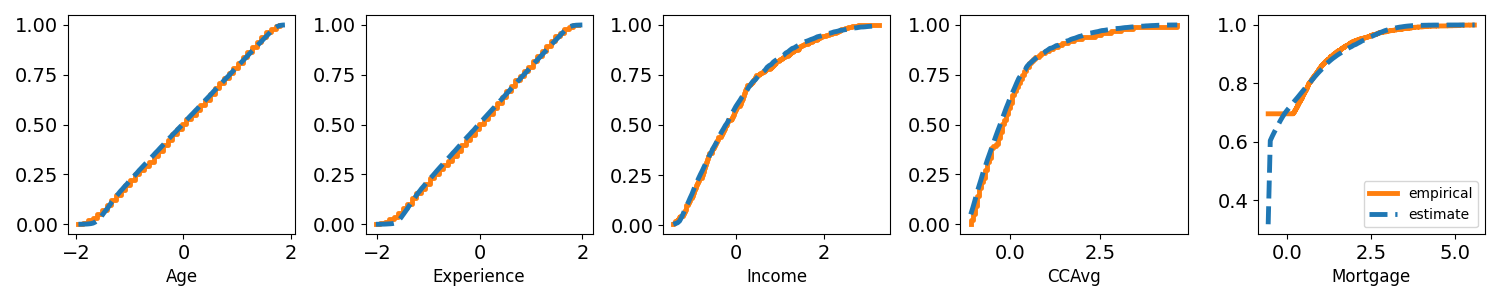}}
    \caption{Empirical and estimated CDFs of continuous and count variables (Monte Carlo approximated with 5000 samples). We standardize covariates and remove observations outside the 1\% and 99\% percentile range.}
\end{figure*}

\begin{figure*}[ht]
    \centering
    \subfigure[\texttt{adult}]{
    \includegraphics[width=\textwidth]{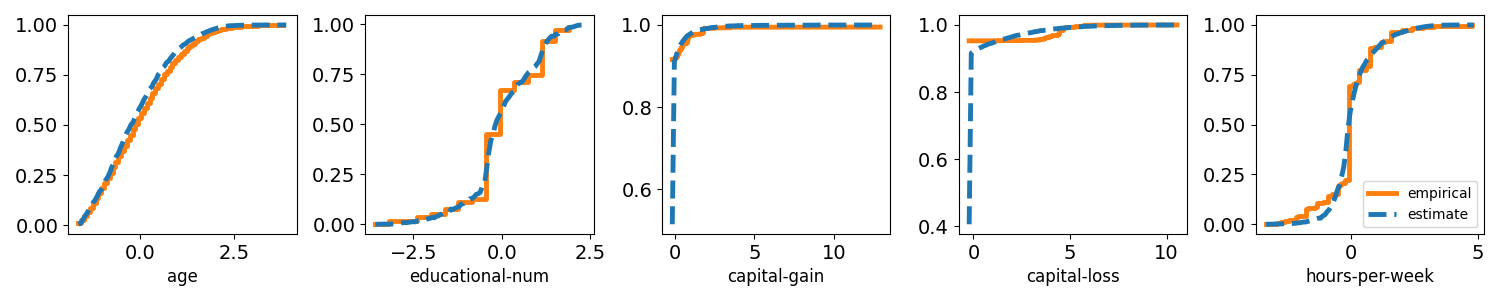}}
    \subfigure[\texttt{cabs}]{
    \includegraphics[width=\textwidth]{fig/cabs_estimated_quantile.png}}
    \subfigure[\texttt{kings}]{
    \includegraphics[width=\textwidth]{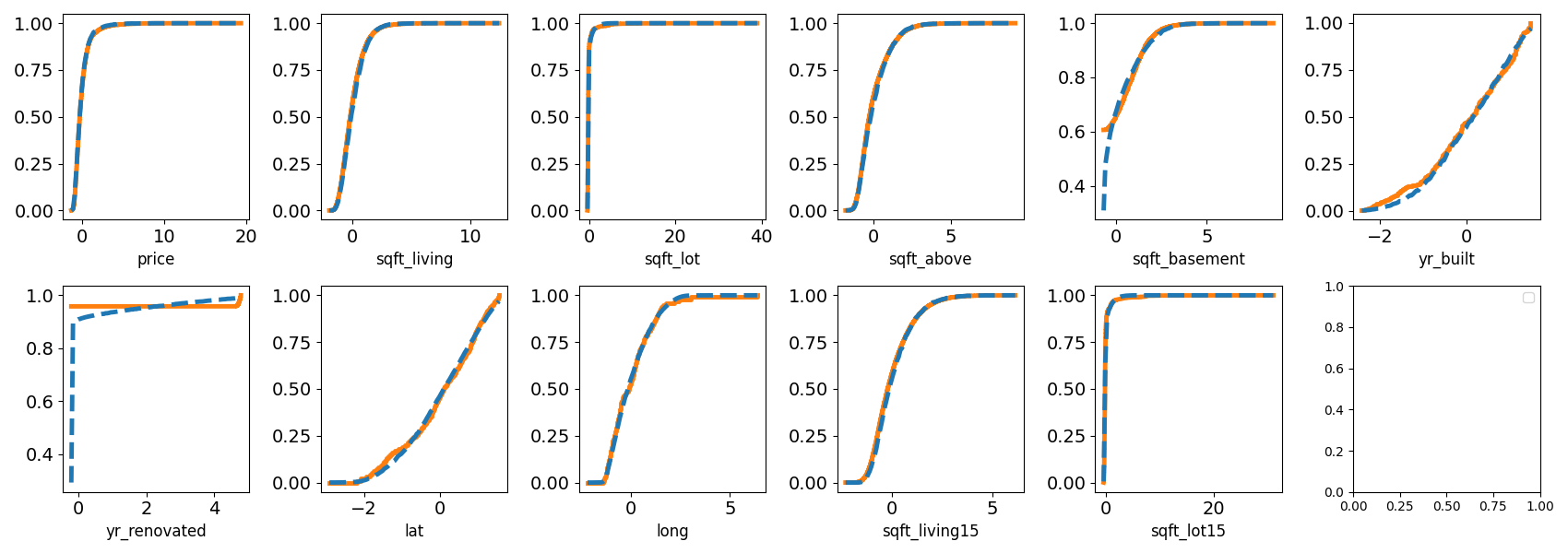}}
    \caption{Empirical and estimated CDFs of continuous and count variables (Monte Carlo approximated with 5000 samples). We standardize covariates and remove observations outside the 1\% and 99\% percentile range.}
\end{figure*}

\clearpage
\begin{table}[ht]
\caption{MLu metrics (MARE, $F_1$). Mean and standard deviation values are obtained from 10 repeated experiments. $\uparrow$ denotes higher is better and $\downarrow$ denotes lower is better.}
\centering
    \resizebox{\textwidth}{!}{\subtable[]{
  \begin{tabular}{lrrrrrr}
    \toprule
    Dataset & \multicolumn{2}{c}{\texttt{covertype}} & \multicolumn{2}{c}{\texttt{credit}} & \multicolumn{2}{c}{\texttt{loan}} \\
    \midrule
    Model & MARE $\downarrow$ & $F_1$ $\uparrow$ & MARE $\downarrow$ & $F_1$ $\uparrow$ & MARE $\downarrow$ & $F_1$ $\uparrow$ \\
    \midrule
Baseline & 0.035 & 0.718 & 0.064 & 0.927 & 0.020 & 0.948\\
CTGAN & $0.058_{\pm 0.007}$ & $0.227_{\pm 0.030}$ & $0.593_{\pm 0.150}$ & $0.914_{\pm 0.006}$ & $0.258_{\pm 0.020}$ & $0.842_{\pm 0.109}$\\
TVAE & $0.079_{\pm 0.007}$ & $0.504_{\pm 0.032}$ & $0.260_{\pm 0.135}$ & $0.091_{\pm 0.286}$ & $0.124_{\pm 0.033}$ & $0.785_{\pm 0.288}$\\
CTAB-GAN & $0.065_{\pm 0.004}$ & $0.493_{\pm 0.027}$ & $0.887_{\pm 0.351}$ & $0.913_{\pm 0.005}$ & $0.247_{\pm 0.019}$ & $0.887_{\pm 0.026}$\\
    \midrule
DistVAE($\beta=0.5$) & $0.044_{\pm 0.002}$ & $0.605_{\pm 0.006}$ & $0.763_{\pm 0.068}$ & $0.926_{\pm 0.001}$ & $0.249_{\pm 0.007}$ & $0.914_{\pm 0.009}$\\
DistVAE($\beta=1$) & $0.045_{\pm 0.001}$ & $0.557_{\pm 0.007}$ & $0.774_{\pm 0.035}$ & $0.926_{\pm 0.000}$ & $0.249_{\pm 0.005}$ & $0.897_{\pm 0.002}$\\
DistVAE($\beta=5$) & $0.063_{\pm 0.001}$ & $0.443_{\pm 0.018}$ & $0.870_{\pm 0.022}$ & $0.904_{\pm 0.010}$ & $0.249_{\pm 0.005}$ & $0.893_{\pm 0.005}$\\
    \bottomrule
  \end{tabular}}}
    
    \resizebox{\textwidth}{!}{\subtable[]{
  \begin{tabular}{lrrrrrr}
    \toprule
    Dataset & \multicolumn{2}{c}{\texttt{adult}} & \multicolumn{2}{c}{\texttt{cabs}} & \multicolumn{2}{c}{\texttt{kings}} \\
    \midrule
    Model & MARE $\downarrow$ & $F_1$ $\uparrow$ & MARE $\downarrow$ & $F_1$ $\uparrow$ & MARE $\downarrow$ & $F_1$ $\uparrow$ \\
    \midrule
Baseline & 0.216 & 0.854 & 0.565 & 0.743 & 0.001 & 0.695\\
CTGAN & $0.297_{\pm 0.030}$ & $0.796_{\pm 0.022}$ & $0.721_{\pm 0.046}$ & $0.674_{\pm 0.024}$ & $0.001_{\pm 0.000}$ & $0.579_{\pm 0.035}$\\
TVAE & $0.238_{\pm 0.006}$ & $0.809_{\pm 0.016}$ & $0.642_{\pm 0.035}$ & $0.689_{\pm 0.031}$ & $0.010_{\pm 0.005}$ & $0.687_{\pm 0.041}$\\
CTAB-GAN & $0.321_{\pm 0.036}$ & $0.730_{\pm 0.069}$ & $0.894_{\pm 0.116}$ & $0.582_{\pm 0.047}$ & $0.001_{\pm 0.000}$ & $0.608_{\pm 0.022}$\\
    \midrule
DistVAE($\beta=0.5$) & $0.232_{\pm 0.004}$ & $0.825_{\pm 0.009}$ & $0.803_{\pm 0.129}$ & $0.707_{\pm 0.010}$ & $0.001_{\pm 0.000}$ & $0.640_{\pm 0.002}$\\
DistVAE($\beta=1$) & $0.234_{\pm 0.006}$ & $0.822_{\pm 0.003}$ & $0.760_{\pm 0.062}$ & $0.725_{\pm 0.004}$ & $0.001_{\pm 0.000}$ & $0.644_{\pm 0.003}$\\
DistVAE($\beta=5$) & $0.327_{\pm 0.008}$ & $0.751_{\pm 0.010}$ & $0.839_{\pm 0.042}$ & $0.447_{\pm 0.009}$ & $0.002_{\pm 0.000}$ & $0.637_{\pm 0.004}$\\
    \bottomrule
  \end{tabular}}}
\label{tab:tabular}
\end{table}

\begin{figure}[ht]
    \centering
    \includegraphics[width=0.6\linewidth]{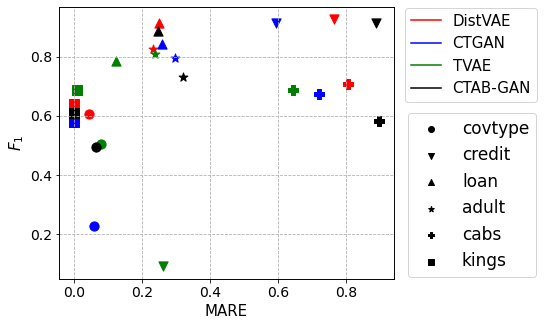}
    \caption{Machine learning utilities for compared models and real tabular datasets.}
    \label{fig:mlu}
\end{figure}

For a detailed comparison of the models and their performance, we present the paired (MARE, $F_1$) scores for all tabular datasets in Figure \ref{fig:mlu}. A better score, indicating a superior performance in terms of MLu, is represented by a dot located in the upper left corner. Notably, Figure \ref{fig:mlu} consistently demonstrates that DistVAE achieves the best or at least competitive MLu across all tabular datasets. TVAE exhibits a notably low $F_1$ score in the \texttt{credit} dataset because it fails to handle the highly imbalanced discrete target variable. This comparative analysis highlights the strong MLu performance of DistVAE and the specific limitations of TVAE in certain scenarios.


\begin{table}[ht]
\caption{Correlation structure similarity. `CorrDist' represents $L_2$ distance between the correlation matrix of synthetic and real datasets. Mean and standard deviation values are obtained from 10 repeated experiments. Lower is better.}
  \centering
  \resizebox{\textwidth}{!}{\begin{tabular}{lrrrrrr}
    \toprule
    Dataset & \texttt{covertype} & \texttt{credit} & \texttt{loan} & \texttt{adult} & \texttt{cabs} & \texttt{kings} \\ 
    \midrule
    Model & CorrDist & CorrDist & CorrDist & CorrDist & CorrDist & CorrDist \\
    \midrule
CTGAN & $2.167_{\pm 0.419}$ & $2.323_{\pm 0.362}$ & $2.282_{\pm 0.177}$ & $1.788_{\pm 0.217}$ & $1.679_{\pm 0.129}$ & $2.839_{\pm 0.246}$\\
TVAE & $1.969_{\pm 0.146}$ & $4.021_{\pm 0.451}$ & $2.404_{\pm 0.408}$ & $2.231_{\pm 0.269}$ & $3.136_{\pm 0.686}$ & $2.665_{\pm 0.296}$\\
CTAB-GAN & $2.351_{\pm 0.185}$ & $2.696_{\pm 0.275}$ & $2.073_{\pm 0.110}$ & $2.387_{\pm 0.470}$ & $2.532_{\pm 0.225}$ & $3.411_{\pm 0.399}$\\
    \midrule
DistVAE($\beta=0.5$) & $1.179_{\pm 0.090}$ & $2.072_{\pm 0.162}$ & $1.654_{\pm 0.050}$ & $0.830_{\pm 0.078}$ & $1.481_{\pm 0.071}$ & $1.559_{\pm 0.135}$\\
DistVAE($\beta=1$) & $2.359_{\pm 0.018}$ & $2.229_{\pm 0.102}$ & $1.910_{\pm 0.019}$ & $0.746_{\pm 0.042}$ & $1.495_{\pm 0.047}$ & $1.621_{\pm 0.149}$\\
DistVAE($\beta=5$) & $2.946_{\pm 0.007}$ & $3.161_{\pm 0.006}$ & $2.113_{\pm 0.015}$ & $3.186_{\pm 0.007}$ & $1.930_{\pm 0.004}$ & $5.339_{\pm 0.006}$\\
    \bottomrule
  \end{tabular}}
\end{table}
\begin{table}[ht]
\caption{Marginal statistical similarity. K-S denotes the Kolmogorov–Smirnov statistic and 1-WD represents the 1-Wasserstein distance. Mean and standard deviation values are obtained from 10 repeated experiments. Lower is better.}
  \centering
  \resizebox{\textwidth}{!}{\subtable[Continuous]{
  \begin{tabular}{lrrrrrr}
    \toprule
    Dataset & \multicolumn{2}{c}{\texttt{covertype}} & \multicolumn{2}{c}{\texttt{credit}} & \multicolumn{2}{c}{\texttt{loan}} \\
    \midrule
    Model & K-S & 1-WD & K-S & 1-WD & K-S & 1-WD \\
    \midrule
CTGAN & $0.080_{\pm 0.011}$ & $0.108_{\pm 0.014}$ & $0.132_{\pm 0.017}$ & $0.088_{\pm 0.013}$ & $0.112_{\pm 0.055}$ & $0.061_{\pm 0.016}$\\
TVAE & $0.088_{\pm 0.008}$ & $0.156_{\pm 0.018}$ & $0.156_{\pm 0.036}$ & $0.200_{\pm 0.035}$ & $0.162_{\pm 0.024}$ & $0.201_{\pm 0.024}$\\
CTAB-GAN & $0.073_{\pm 0.011}$ & $0.096_{\pm 0.014}$ & $0.140_{\pm 0.021}$ & $0.111_{\pm 0.013}$ & $0.181_{\pm 0.070}$ & $0.153_{\pm 0.023}$\\
    \midrule
DistVAE($\beta=0.5$) & $0.032_{\pm 0.003}$ & $0.041_{\pm 0.005}$ & $0.088_{\pm 0.014}$ & $0.089_{\pm 0.004}$ & $0.060_{\pm 0.010}$ & $0.048_{\pm 0.003}$\\
DistVAE($\beta=1$) & $0.049_{\pm 0.001}$ & $0.079_{\pm 0.001}$ & $0.084_{\pm 0.008}$ & $0.093_{\pm 0.002}$ & $0.056_{\pm 0.003}$ & $0.054_{\pm 0.002}$\\
DistVAE($\beta=5$) & $0.064_{\pm 0.001}$ & $0.112_{\pm 0.000}$ & $0.146_{\pm 0.003}$ & $0.113_{\pm 0.001}$ & $0.057_{\pm 0.003}$ & $0.056_{\pm 0.002}$\\
    \bottomrule
  \end{tabular}}}
  
  \resizebox{\textwidth}{!}{\subtable[Continuous]{
  \begin{tabular}{lrrrrrr}
    \toprule
    Dataset & \multicolumn{2}{c}{\texttt{adult}} & \multicolumn{2}{c}{\texttt{cabs}} & \multicolumn{2}{c}{\texttt{kings}} \\
    \midrule
    Model & K-S & 1-WD & K-S & 1-WD & K-S & 1-WD \\
    \midrule
CTGAN & $0.323_{\pm 0.126}$ & $0.086_{\pm 0.017}$ & $0.045_{\pm 0.007}$ & $0.060_{\pm 0.011}$ & $0.109_{\pm 0.012}$ & $0.116_{\pm 0.016}$\\
TVAE & $0.477_{\pm 0.053}$ & $0.414_{\pm 0.070}$ & $0.098_{\pm 0.008}$ & $0.139_{\pm 0.023}$ & $0.195_{\pm 0.018}$ & $0.213_{\pm 0.046}$\\
CTAB-GAN & $0.275_{\pm 0.122}$ & $0.178_{\pm 0.053}$ & $0.086_{\pm 0.008}$ & $0.118_{\pm 0.015}$ & $0.188_{\pm 0.035}$ & $0.128_{\pm 0.019}$\\
    \midrule
DistVAE($\beta=0.5$) & $0.209_{\pm 0.064}$ & $0.114_{\pm 0.009}$ & $0.044_{\pm 0.003}$ & $0.067_{\pm 0.003}$ & $0.110_{\pm 0.008}$ & $0.089_{\pm 0.003}$\\
DistVAE($\beta=1$) & $0.138_{\pm 0.041}$ & $0.111_{\pm 0.007}$ & $0.046_{\pm 0.002}$ & $0.068_{\pm 0.003}$ & $0.115_{\pm 0.007}$ & $0.093_{\pm 0.002}$\\
DistVAE($\beta=5$) & $0.115_{\pm 0.005}$ & $0.234_{\pm 0.002}$ & $0.052_{\pm 0.001}$ & $0.074_{\pm 0.002}$ & $0.120_{\pm 0.002}$ & $0.137_{\pm 0.002}$\\
    \bottomrule
  \end{tabular}}}

  \resizebox{\textwidth}{!}{\subtable[Discrete]{
  \begin{tabular}{lrrrrrr}
    \toprule
    Dataset & \multicolumn{2}{c}{\texttt{covertype}} & \multicolumn{2}{c}{\texttt{credit}} & \multicolumn{2}{c}{\texttt{loan}} \\
    \midrule
    Model & K-S & 1-WD & K-S & 1-WD & K-S & 1-WD \\
    \midrule
CTGAN & $0.591_{\pm 0.003}$ & $1.629_{\pm 0.011}$ & $0.061_{\pm 0.008}$ & $0.147_{\pm 0.024}$ & $0.070_{\pm 0.010}$ & $0.076_{\pm 0.013}$\\
TVAE & $0.238_{\pm 0.042}$ & $0.606_{\pm 0.033}$ & $0.583_{\pm 0.045}$ & $1.566_{\pm 0.116}$ & $0.193_{\pm 0.028}$ & $0.221_{\pm 0.043}$\\
CTAB-GAN & $0.052_{\pm 0.028}$ & $0.180_{\pm 0.126}$ & $0.034_{\pm 0.006}$ & $0.076_{\pm 0.018}$ & $0.039_{\pm 0.011}$ & $0.046_{\pm 0.013}$\\
    \midrule
DistVAE($\beta=0.5$) & $0.023_{\pm 0.010}$ & $0.073_{\pm 0.046}$ & $0.020_{\pm 0.003}$ & $0.046_{\pm 0.006}$ & $0.019_{\pm 0.005}$ & $0.027_{\pm 0.008}$\\
DistVAE($\beta=1$) & $0.011_{\pm 0.005}$ & $0.036_{\pm 0.025}$ & $0.018_{\pm 0.001}$ & $0.042_{\pm 0.003}$ & $0.011_{\pm 0.002}$ & $0.015_{\pm 0.005}$\\
DistVAE($\beta=5$) & $0.109_{\pm 0.003}$ & $0.379_{\pm 0.014}$ & $0.027_{\pm 0.001}$ & $0.068_{\pm 0.004}$ & $0.009_{\pm 0.002}$ & $0.013_{\pm 0.005}$\\
    \bottomrule
  \end{tabular}}}
  
  \resizebox{\textwidth}{!}{\subtable[Discrete]{
  \begin{tabular}{lrrrrrr}
    \toprule
    Dataset & \multicolumn{2}{c}{\texttt{adult}} & \multicolumn{2}{c}{\texttt{cabs}} & \multicolumn{2}{c}{\texttt{kings}} \\
    \midrule
    Model & K-S & 1-WD & K-S & 1-WD & K-S & 1-WD \\
    \midrule
CTGAN & $0.065_{\pm 0.008}$ & $0.463_{\pm 0.051}$ & $0.069_{\pm 0.016}$ & $0.238_{\pm 0.036}$ & $0.140_{\pm 0.018}$ & $0.529_{\pm 0.064}$\\
TVAE & $0.479_{\pm 0.048}$ & $5.228_{\pm 0.273}$ & $0.411_{\pm 0.088}$ & $1.202_{\pm 0.212}$ & $0.405_{\pm 0.034}$ & $1.262_{\pm 0.099}$\\
CTAB-GAN & $0.169_{\pm 0.018}$ & $0.745_{\pm 0.294}$ & $0.086_{\pm 0.013}$ & $0.443_{\pm 0.074}$ & $0.255_{\pm 0.027}$ & $0.979_{\pm 0.107}$\\
    \midrule
DistVAE($\beta=0.5$) & $0.037_{\pm 0.004}$ & $0.248_{\pm 0.026}$ & $0.060_{\pm 0.017}$ & $0.241_{\pm 0.084}$ & $0.022_{\pm 0.004}$ & $0.071_{\pm 0.012}$\\
DistVAE($\beta=1$) & $0.031_{\pm 0.004}$ & $0.239_{\pm 0.019}$ & $0.070_{\pm 0.003}$ & $0.292_{\pm 0.010}$ & $0.023_{\pm 0.003}$ & $0.073_{\pm 0.008}$\\
DistVAE($\beta=5$) & $0.070_{\pm 0.002}$ & $0.457_{\pm 0.010}$ & $0.083_{\pm 0.003}$ & $0.326_{\pm 0.011}$ & $0.059_{\pm 0.002}$ & $0.204_{\pm 0.011}$\\
    \bottomrule
  \end{tabular}}}
\end{table}
\begin{table*}[ht]
\caption{Privacy preservability: Distance to closest record (DCR) between real training and synthetic datasets (R\&S), between the same real training datasets (R), and between the same synthetic datasets (S). Mean and standard deviation values are obtained from 10 repeated experiments. Higher is better.}
\centering
\subtable[]{
  \resizebox{0.49\textwidth}{!}{\begin{tabular}{lrrr}
    \toprule
    Dataset & \multicolumn{3}{c}{\texttt{covertype}} \\
    \midrule
    Model & R\&S & R & S \\
    \midrule
CTGAN & $0.715_{\pm 0.026}$ & $0.329_{\pm 0.000}$ & $0.514_{\pm 0.094}$\\
TVAE & $0.676_{\pm 0.031}$ & $0.329_{\pm 0.000}$ & $0.482_{\pm 0.025}$\\
CTAB-GAN & $0.892_{\pm 0.031}$ & $0.329_{\pm 0.000}$ & $0.011_{\pm 0.005}$\\
\midrule
DistVAE($\beta=0.5$) & $0.765_{\pm 0.008}$ & $0.329_{\pm 0.000}$ & $0.819_{\pm 0.010}$\\
DistVAE($\beta=1$) & $0.878_{\pm 0.008}$ & $0.329_{\pm 0.000}$ & $0.906_{\pm 0.009}$\\
DistVAE($\beta=5$) & $0.907_{\pm 0.012}$ & $0.329_{\pm 0.000}$ & $0.939_{\pm 0.008}$\\
    \bottomrule
  \end{tabular}
}}
\subtable[]{
  \resizebox{0.49\textwidth}{!}{\begin{tabular}{lrrr}
    \toprule
    Dataset & \multicolumn{3}{c}{\texttt{credit}} \\
    \midrule
    Model & R\&S & R & S \\
    \midrule
CTGAN & $0.624_{\pm 0.033}$ & $0.452_{\pm 0.000}$ & $0.592_{\pm 0.061}$\\
TVAE & $0.627_{\pm 0.118}$ & $0.452_{\pm 0.000}$ & $0.423_{\pm 0.212}$\\
CTAB-GAN & $0.715_{\pm 0.025}$ & $0.452_{\pm 0.000}$ & $0.014_{\pm 0.006}$\\
\midrule
DistVAE($\beta=0.5$) & $0.692_{\pm 0.012}$ & $0.452_{\pm 0.000}$ & $0.742_{\pm 0.012}$\\
DistVAE($\beta=1$) & $0.700_{\pm 0.006}$ & $0.452_{\pm 0.000}$ & $0.750_{\pm 0.005}$\\
DistVAE($\beta=5$) & $0.718_{\pm 0.007}$ & $0.452_{\pm 0.000}$ & $0.757_{\pm 0.007}$\\
    \bottomrule
  \end{tabular}
}}
\subtable[]{
  \resizebox{0.49\textwidth}{!}{\begin{tabular}{lrrr}
    \toprule
    Dataset & \multicolumn{3}{c}{\texttt{loan}} \\
    \midrule
    Model & R\&S & R & S \\
    \midrule
CTGAN & $0.249_{\pm 0.017}$ & $0.109_{\pm 0.000}$ & $0.243_{\pm 0.024}$\\
TVAE & $0.298_{\pm 0.057}$ & $0.109_{\pm 0.000}$ & $0.154_{\pm 0.026}$\\
CTAB-GAN & $0.272_{\pm 0.024}$ & $0.109_{\pm 0.000}$ & $0.076_{\pm 0.051}$\\
\midrule
DistVAE($\beta=0.5$) & $0.244_{\pm 0.025}$ & $0.109_{\pm 0.000}$ & $0.245_{\pm 0.020}$\\
DistVAE($\beta=1$) & $0.238_{\pm 0.019}$ & $0.109_{\pm 0.000}$ & $0.241_{\pm 0.023}$\\
DistVAE($\beta=5$) & $0.243_{\pm 0.015}$ & $0.109_{\pm 0.000}$ & $0.240_{\pm 0.025}$\\
    \bottomrule
  \end{tabular}
}}
\subtable[]{
  \resizebox{0.49\textwidth}{!}{\begin{tabular}{lrrr}
    \toprule
    Dataset & \multicolumn{3}{c}{\texttt{adult}} \\
    \midrule
    Model & R\&S & R & S \\
    \midrule
CTGAN & $0.063_{\pm 0.017}$ & $0.000_{\pm 0.000}$ & $0.000_{\pm 0.000}$\\
TVAE & $0.277_{\pm 0.039}$ & $0.000_{\pm 0.000}$ & $0.000_{\pm 0.000}$\\
CTAB-GAN & $0.152_{\pm 0.064}$ & $0.000_{\pm 0.000}$ & $0.000_{\pm 0.000}$\\
\midrule
DistVAE($\beta=0.5$) & $0.060_{\pm 0.040}$ & $0.000_{\pm 0.000}$ & $0.005_{\pm 0.001}$\\
DistVAE($\beta=1$) & $0.048_{\pm 0.027}$ & $0.000_{\pm 0.000}$ & $0.003_{\pm 0.000}$\\
DistVAE($\beta=5$) & $0.177_{\pm 0.002}$ & $0.000_{\pm 0.000}$ & $0.001_{\pm 0.000}$\\
    \bottomrule
  \end{tabular}
}}
\subtable[]{
  \resizebox{0.49\textwidth}{!}{\begin{tabular}{lrrr}
    \toprule
    Dataset & \multicolumn{3}{c}{\texttt{cabs}} \\
    \midrule
    Model & R\&S & R & S \\
    \midrule
CTGAN & $0.353_{\pm 0.006}$ & $0.332_{\pm 0.000}$ & $0.341_{\pm 0.010}$\\
TVAE & $0.339_{\pm 0.008}$ & $0.332_{\pm 0.000}$ & $0.195_{\pm 0.015}$\\
CTAB-GAN & $0.423_{\pm 0.025}$ & $0.332_{\pm 0.000}$ & $0.012_{\pm 0.023}$\\
\midrule
DistVAE($\beta=0.5$) & $0.364_{\pm 0.004}$ & $0.332_{\pm 0.000}$ & $0.368_{\pm 0.005}$\\
DistVAE($\beta=1$) & $0.364_{\pm 0.004}$ & $0.332_{\pm 0.000}$ & $0.367_{\pm 0.004}$\\
DistVAE($\beta=5$) & $0.368_{\pm 0.003}$ & $0.332_{\pm 0.000}$ & $0.365_{\pm 0.005}$\\
    \bottomrule
  \end{tabular}
}}
\subtable[]{
  \resizebox{0.49\textwidth}{!}{\begin{tabular}{lrrr}
    \toprule
    Dataset & \multicolumn{3}{c}{\texttt{kings}} \\
    \midrule
    Model & R\&S & R & S \\
    \midrule
CTGAN & $0.550_{\pm 0.016}$ & $0.199_{\pm 0.000}$ & $0.447_{\pm 0.030}$\\
TVAE & $0.603_{\pm 0.091}$ & $0.199_{\pm 0.000}$ & $0.414_{\pm 0.053}$\\
CTAB-GAN & $0.596_{\pm 0.030}$ & $0.199_{\pm 0.000}$ & $0.122_{\pm 0.135}$\\
\midrule
DistVAE($\beta=0.5$) & $0.540_{\pm 0.009}$ & $0.199_{\pm 0.000}$ & $0.600_{\pm 0.013}$\\
DistVAE($\beta=1$) & $0.552_{\pm 0.008}$ & $0.199_{\pm 0.000}$ & $0.605_{\pm 0.014}$\\
DistVAE($\beta=5$) & $0.692_{\pm 0.013}$ & $0.199_{\pm 0.000}$ & $0.763_{\pm 0.015}$\\
    \bottomrule
  \end{tabular}
}}
\end{table*}
\begin{table*}[ht]
\caption{Privacy preservability: Membership inference attack performance. Mean and standard deviation values are obtained from 10 repeated experiments.}
\centering
\subtable[]{
  \resizebox{0.45\textwidth}{!}{\begin{tabular}{lrr}
    \toprule
    Dataset & \multicolumn{2}{c}{\texttt{covertype}} \\
    \midrule
    Model & Accuracy & AUC \\
    \midrule
TVAE & $0.499_{\pm 0.007}$ & $0.499_{\pm 0.007}$\\
DistVAE($\beta=0.5$) & $0.500_{\pm 0.003}$ & $0.500_{\pm 0.003}$\\
    \bottomrule
  \end{tabular}
}}
\subtable[]{
  \resizebox{0.45\textwidth}{!}{\begin{tabular}{lrr}
    \toprule
    Dataset & \multicolumn{2}{c}{\texttt{credit}} \\
    \midrule
    Model & Accuracy & AUC \\
    \midrule
TVAE & $0.500_{\pm 0.001}$ & $0.500_{\pm 0.001}$\\
DistVAE($\beta=0.5$) & $0.500_{\pm 0.001}$ & $0.500_{\pm 0.001}$\\
    \bottomrule
  \end{tabular}
}}
\subtable[]{
  \resizebox{0.45\textwidth}{!}{\begin{tabular}{lrr}
    \toprule
    Dataset & \multicolumn{2}{c}{\texttt{loan}} \\
    \midrule
    Model & Accuracy & AUC \\
    \midrule
TVAE & $0.497_{\pm 0.015}$ & $0.497_{\pm 0.015}$\\
DistVAE($\beta=0.5$) & $0.502_{\pm 0.006}$ & $0.502_{\pm 0.006}$\\
    \bottomrule
  \end{tabular}
}}
\subtable[]{
  \resizebox{0.45\textwidth}{!}{\begin{tabular}{lrr}
    \toprule
    Dataset & \multicolumn{2}{c}{\texttt{adult}} \\
    \midrule
    Model & Accuracy & AUC \\
    \midrule
TVAE & $0.493_{\pm 0.017}$ & $0.493_{\pm 0.017}$\\
DistVAE($\beta=0.5$) & $0.500_{\pm 0.000}$ & $0.500_{\pm 0.000}$\\
    \bottomrule
  \end{tabular}
}}
\subtable[]{
  \resizebox{0.45\textwidth}{!}{\begin{tabular}{lrr}
    \toprule
    Dataset & \multicolumn{2}{c}{\texttt{cabs}} \\
    \midrule
    Model & Accuracy & AUC \\
    \midrule
TVAE & $0.480_{\pm 0.033}$ & $0.480_{\pm 0.033}$\\
DistVAE($\beta=0.5$) & $0.498_{\pm 0.003}$ & $0.498_{\pm 0.003}$\\
    \bottomrule
  \end{tabular}
}}
\subtable[]{
  \resizebox{0.45\textwidth}{!}{\begin{tabular}{lrr}
    \toprule
    Dataset & \multicolumn{2}{c}{\texttt{kings}} \\
    \midrule
    Model & Accuracy & AUC \\
    \midrule
TVAE & $0.507_{\pm 0.025}$ & $0.507_{\pm 0.025}$\\
DistVAE($\beta=0.5$) & $0.502_{\pm 0.004}$ & $0.502_{\pm 0.004}$\\
    \bottomrule
  \end{tabular}
}}
\end{table*}
\begin{table*}[ht]
\caption{Privacy preservability: Attribute disclosure performance with $F_1$ score. Mean and standard deviation values are obtained from 10 repeated experiments. Lower is better.}
\centering
\subtable[\texttt{covtype}]{
  \resizebox{0.49\textwidth}{!}{\begin{tabular}{lrrr}
    \toprule
    & \multicolumn{3}{c}{Number of neighbors ($k$)} \\
    \cmidrule(){2-4}
    Model & 1 & 10 & 100 \\
    \midrule
CTGAN & $0.161_{\pm 0.030}$ & $0.175_{\pm 0.029}$ & $0.155_{\pm 0.041}$\\
TVAE & $0.356_{\pm 0.086}$ & $0.357_{\pm 0.091}$ & $0.349_{\pm 0.094}$\\
CTAB-GAN & $0.200_{\pm 0.035}$ & $0.225_{\pm 0.042}$ & $0.238_{\pm 0.033}$\\
\midrule
DistVAE($\beta=0.5$) & $0.308_{\pm 0.025}$ & $0.338_{\pm 0.037}$ & $0.313_{\pm 0.035}$\\
DistVAE($\beta=1$) & $0.264_{\pm 0.024}$ & $0.294_{\pm 0.036}$ & $0.282_{\pm 0.029}$\\
DistVAE($\beta=5$) & $0.141_{\pm 0.019}$ & $0.135_{\pm 0.013}$ & $0.115_{\pm 0.011}$\\
    \bottomrule
  \end{tabular}
}}
\subtable[\texttt{credit}]{
  \resizebox{0.49\textwidth}{!}{\begin{tabular}{lrrr}
    \toprule
    & \multicolumn{3}{c}{Number of neighbors ($k$)} \\
    \cmidrule(){2-4}
    Model & 1 & 10 & 100 \\
    \midrule
CTGAN & $0.319_{\pm 0.009}$ & $0.330_{\pm 0.009}$ & $0.323_{\pm 0.011}$\\
TVAE & $0.615_{\pm 0.147}$ & $0.618_{\pm 0.143}$ & $0.614_{\pm 0.145}$\\
CTAB-GAN & $0.322_{\pm 0.007}$ & $0.332_{\pm 0.012}$ & $0.331_{\pm 0.010}$\\
\midrule
DistVAE($\beta=0.5$) & $0.339_{\pm 0.013}$ & $0.337_{\pm 0.011}$ & $0.315_{\pm 0.012}$\\
DistVAE($\beta=1$) & $0.339_{\pm 0.010}$ & $0.324_{\pm 0.005}$ & $0.301_{\pm 0.004}$\\
DistVAE($\beta=5$) & $0.332_{\pm 0.007}$ & $0.317_{\pm 0.007}$ & $0.293_{\pm 0.007}$\\
    \bottomrule
  \end{tabular}
}}
\subtable[\texttt{loan}]{
  \resizebox{0.49\textwidth}{!}{\begin{tabular}{lrrr}
    \toprule
    & \multicolumn{3}{c}{Number of neighbors ($k$)} \\
    \cmidrule(){2-4}
    Model & 1 & 10 & 100 \\
    \midrule
CTGAN & $0.439_{\pm 0.027}$ & $0.447_{\pm 0.022}$ & $0.426_{\pm 0.033}$\\
TVAE & $0.611_{\pm 0.153}$ & $0.602_{\pm 0.152}$ & $0.596_{\pm 0.157}$\\
CTAB-GAN & $0.475_{\pm 0.048}$ & $0.443_{\pm 0.027}$ & $0.435_{\pm 0.030}$\\
\midrule
DistVAE($\beta=0.5$) & $0.505_{\pm 0.043}$ & $0.465_{\pm 0.038}$ & $0.439_{\pm 0.032}$\\
DistVAE($\beta=1$) & $0.449_{\pm 0.038}$ & $0.440_{\pm 0.035}$ & $0.423_{\pm 0.031}$\\
DistVAE($\beta=5$) & $0.458_{\pm 0.013}$ & $0.441_{\pm 0.035}$ & $0.416_{\pm 0.031}$\\
    \bottomrule
  \end{tabular}
}}
\subtable[\texttt{adult}]{
  \resizebox{0.49\textwidth}{!}{\begin{tabular}{lrrr}
    \toprule
    & \multicolumn{3}{c}{Number of neighbors ($k$)} \\
    \cmidrule(){2-4}
    Model & 1 & 10 & 100 \\
    \midrule
CTGAN & $0.234_{\pm 0.012}$ & $0.255_{\pm 0.021}$ & $0.261_{\pm 0.023}$\\
TVAE & $0.318_{\pm 0.081}$ & $0.318_{\pm 0.080}$ & $0.307_{\pm 0.080}$\\
CTAB-GAN & $0.199_{\pm 0.032}$ & $0.202_{\pm 0.032}$ & $0.205_{\pm 0.031}$\\
\midrule
DistVAE($\beta=0.5$) & $0.270_{\pm 0.018}$ & $0.280_{\pm 0.021}$ & $0.268_{\pm 0.015}$\\
DistVAE($\beta=1$) & $0.264_{\pm 0.006}$ & $0.276_{\pm 0.005}$ & $0.260_{\pm 0.007}$\\
DistVAE($\beta=5$) & $0.205_{\pm 0.005}$ & $0.187_{\pm 0.003}$ & $0.166_{\pm 0.002}$\\
    \bottomrule
  \end{tabular}
}}
\subtable[\texttt{cabs}]{
  \resizebox{0.49\textwidth}{!}{\begin{tabular}{lrrr}
    \toprule
    & \multicolumn{3}{c}{Number of neighbors ($k$)} \\
    \cmidrule(){2-4}
    Model & 1 & 10 & 100 \\
    \midrule
CTGAN & $0.238_{\pm 0.009}$ & $0.246_{\pm 0.006}$ & $0.231_{\pm 0.010}$\\
TVAE & $0.385_{\pm 0.079}$ & $0.383_{\pm 0.079}$ & $0.381_{\pm 0.081}$\\
CTAB-GAN & $0.235_{\pm 0.010}$ & $0.236_{\pm 0.008}$ & $0.237_{\pm 0.011}$\\
\midrule
DistVAE($\beta=0.5$) & $0.251_{\pm 0.010}$ & $0.241_{\pm 0.007}$ & $0.225_{\pm 0.005}$\\
DistVAE($\beta=1$) & $0.246_{\pm 0.010}$ & $0.240_{\pm 0.006}$ & $0.227_{\pm 0.008}$\\
DistVAE($\beta=5$) & $0.238_{\pm 0.010}$ & $0.220_{\pm 0.009}$ & $0.199_{\pm 0.005}$\\
    \bottomrule
  \end{tabular}
}}
\subtable[\texttt{kings}]{
  \resizebox{0.49\textwidth}{!}{\begin{tabular}{lrrr}
    \toprule
    & \multicolumn{3}{c}{Number of neighbors ($k$)} \\
    \cmidrule(){2-4}
    Model & 1 & 10 & 100 \\
    \midrule
CTGAN & $0.200_{\pm 0.009}$ & $0.257_{\pm 0.032}$ & $0.269_{\pm 0.036}$\\
TVAE & $0.338_{\pm 0.045}$ & $0.349_{\pm 0.046}$ & $0.345_{\pm 0.044}$\\
CTAB-GAN & $0.111_{\pm 0.068}$ & $0.111_{\pm 0.088}$ & $0.123_{\pm 0.114}$\\
\midrule
DistVAE($\beta=0.5$) & $0.293_{\pm 0.019}$ & $0.310_{\pm 0.041}$ & $0.301_{\pm 0.050}$\\
DistVAE($\beta=1$) & $0.281_{\pm 0.016}$ & $0.306_{\pm 0.039}$ & $0.290_{\pm 0.041}$\\
DistVAE($\beta=5$) & $0.215_{\pm 0.010}$ & $0.221_{\pm 0.036}$ & $0.205_{\pm 0.038}$\\
    \bottomrule
  \end{tabular}
}}
\end{table*}



\end{document}